\pdfoutput=1

\documentclass[11pt]{article}

\usepackage{naacl2021}

\usepackage{times}
\usepackage{latexsym}

\usepackage[T1]{fontenc}

\usepackage[utf8]{inputenc}

\usepackage{booktabs}
\usepackage{graphicx}
\usepackage{lipsum}
\usepackage{xspace}
\usepackage{multicol}
\usepackage{mathrsfs}
\usepackage{multirow}
\usepackage{caption}
\usepackage{amssymb}
\usepackage{enumitem}
\usepackage{changepage}
\usepackage{amsmath}
\usepackage{tabularx}
\usepackage{comment}
\usepackage{subcaption}
\usepackage{pdflscape}

\usepackage{textcomp}
\usepackage{url}

\usepackage{microtype}
\usepackage{array}
\newcolumntype{C}[1]{>{\centering}m{#1}}

\newcommand{\onetoone}{\texttt{One2One}\xspace}
\newcommand{\onetoseq}{\texttt{One2Seq}\xspace}

\newcommand{\pk}{\textbf{P@$k$}\xspace}
\newcommand{\rk}{\textbf{R@$k$}\xspace}
\newcommand{\fk}{\textbf{F$_1$@$k$}\xspace}
\newcommand{\fm}{\textbf{F$_1$@$\mathcal{M}$}\xspace}
\newcommand{\ften}{\textbf{F$_1$@10}\xspace}
\newcommand{\ffive}{\textbf{F$_1$@5}\xspace}
\newcommand{\fo}{\textbf{F$_1$@$\mathcal{O}$}\xspace}
\newcommand{\rfifty}{\textbf{R@50}\xspace}
\newcommand{\rten}{\textbf{R@10}\xspace}

\newcommand{\nofm}{\textbf{$\mathcal{M}$}\xspace}
\newcommand{\noften}{\textbf{10}\xspace}
\newcommand{\noffive}{\textbf{5}\xspace}
\newcommand{\nofo}{\textbf{$\mathcal{O}$}\xspace}
\newcommand{\norfifty}{\textbf{50}\xspace}
\newcommand{\norten}{\textbf{10}\xspace}
\newcommand{\noratm}{\textbf{$\mathcal{M}$}\xspace}

\newcommand{\bigrnn}{\texttt{\textsc{BigRNN}}\xspace}
\newcommand{\rnn}{\texttt{\textsc{RNN}}\xspace}
\newcommand{\transformer}{\texttt{\textsc{TRANS}}\xspace}
\newcommand{\rnnstrip}{\texttt{\textsc{RNN}}}
\newcommand{\transformerstrip}{\texttt{\textsc{TRANS}}}

\newcommand{\greedy}{\texttt{Greedy}}  

\definecolor{newcolor1}{HTML}{8005A5}
\definecolor{newcolor2}{HTML}{A50520}
\definecolor{newcolor3}{HTML}{ff830f}
\definecolor{newcolor4}{HTML}{028792}
\definecolor{newcolor5}{HTML}{257101}
\definecolor{newcolor6}{HTML}{854800}

\newcommand{\ori}{\textcolor{newcolor1}{\textbf{\texttt{\textsc{Ori}}}}\xspace}
\newcommand{\orirev}{\textcolor{newcolor1}{\textbf{\texttt{\textsc{Ori-Rev}}}}\xspace}
\newcommand{\random}{\textcolor{newcolor2}{\textbf{\texttt{\textsc{Random}}}}\xspace}
\newcommand{\alphab}{\textcolor{newcolor3}{\textbf{\texttt{\textsc{Alpha}}}}\xspace}
\newcommand{\alpharev}{\textcolor{newcolor3}{\textbf{\texttt{\textsc{Alpha-Rev}}}}\xspace}
\newcommand{\stol}{\textcolor{newcolor4}{\textbf{\texttt{\textsc{S-->L}}}}\xspace}
\newcommand{\ltos}{\textcolor{newcolor4}{\textbf{\texttt{\textsc{L-->S}}}}\xspace}
\newcommand{\abspres}{\textcolor{newcolor5}{\textbf{\texttt{\textsc{Abs-Pres}}}}\xspace}
\newcommand{\presabs}{\textcolor{newcolor6}{\textbf{\texttt{\textsc{Pres-Abs}}}}\xspace}

\newcommand{\onlymix}{\textcolor{newcolor1}{\textsc{Only}\xspace}}
\newcommand{\ftmix}{\textcolor{newcolor2}{\textsc{FT}\xspace}}
\newcommand{\mxmix}{\textcolor{newcolor3}{\textsc{MX}\xspace}}
\newcommand{\altmix}{\textcolor{newcolor4}{\textsc{ALT}\xspace}}

\newcommand{\kpk}{\texttt{\textsc{KP20k}}\xspace}
\newcommand{\kpktrain}{\texttt{\textsc{KP20k}-train}\xspace}
\newcommand{\kpkvalid}{\texttt{\textsc{KP20k-valid}}\xspace}
\newcommand{\kpktest}{\texttt{\textsc{KP20k-test}}\xspace}
\newcommand{\magkp}{\texttt{\textsc{MAGKP}}\xspace}
\newcommand{\magkpln}{\texttt{\textsc{MAGKP}-LN}\xspace}
\newcommand{\magkpnlarge}{\texttt{\textsc{MAGKP}-Nlarge}\xspace}
\newcommand{\magkpnsmall}{\texttt{\textsc{MAGKP}-Nsmall}\xspace}

\newcommand{\duc}{\texttt{\textsc{DUC}}\xspace}
\newcommand{\inspec}{\texttt{\textsc{Inspec}}\xspace}
\newcommand{\krapivin}{\texttt{\textsc{Krapivin}}\xspace}
\newcommand{\nus}{\texttt{\textsc{NUS}}\xspace}
\newcommand{\semeval}{\texttt{\textsc{SemEval}}\xspace}

\newcommand{\bos}{\texttt{<bos>}\xspace}
\newcommand{\sep}{\texttt{<sep>}\xspace}
\newcommand{\eos}{\texttt{<eos>}\xspace}

\title{An Empirical Study on Neural Keyphrase Generation}

\author{Rui Meng$^\dag$ \:\:\:\: Xingdi Yuan$^\ddag$ \:\:\:\: Tong Wang$^{\ddag}$ \:\:\:\: Sanqiang Zhao$^\dag$ \\ 
\textbf{Adam Trischler$^{\ddag}$ \:\:\:\: Daqing He$^{\dag}$}\\
$^\dag$School of Computing and Information, University of Pittsburgh \\
$^\ddag$Microsoft Research, Montr\'{e}al \\
rui.meng@pitt.edu
}

\begin{document}
\maketitle
\begin{abstract}
Recent years have seen a flourishing of neural keyphrase generation (KPG) works, including the release of several large-scale datasets and a host of new models to tackle them.
Model performance on KPG tasks has increased significantly with evolving deep learning research.
However, there lacks a comprehensive comparison among different model designs, and a thorough investigation on related factors that may affect a KPG system's generalization performance.
In this empirical study, we aim to fill this gap by providing extensive experimental results and analyzing the most crucial factors impacting the generalizability of KPG models.
We hope this study can help clarify some of the uncertainties surrounding the KPG task and facilitate future research on this topic.
\end{abstract}

\section{Introduction}
\label{section:intro}

Keyphrases are phrases that summarize and highlight important information in a piece of text.
Keyphrase generation (KPG) is the task of automatically predicting such keyphrases given the source text.
The task can be (and has often been) easily misunderstood and trivialized as yet another natural language generation task like summarization and translation, failing to recognize one key aspect that distinguishes KPG: the multiplicity of generation targets; for each input sequence, a KPG system is expected to output \emph{multiple} keyphrases, each a mini-sequence of multiple word tokens.



\begin{figure*}[t!]
    \centering
    \includegraphics[width=1.0\textwidth]{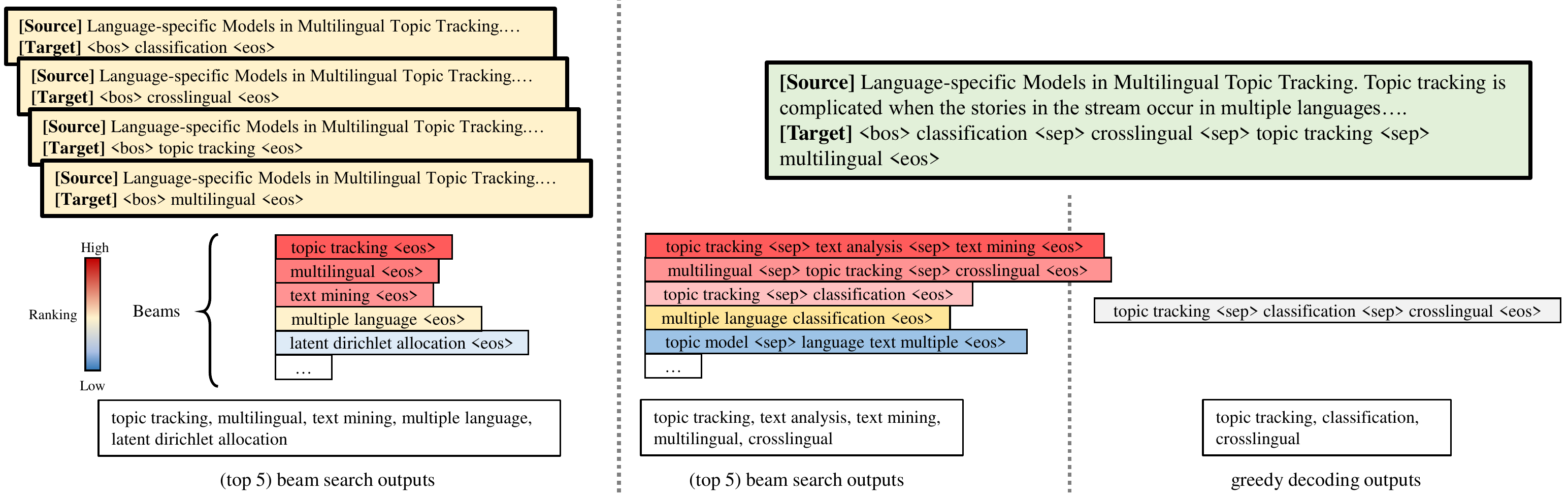}
    \caption{\textbf{Top:} comparison between \onetoone (left) and \onetoseq (right) paradigms on the same data point. \textbf{Bottom:} demonstration of the decoding process for \onetoone (left) and \onetoseq (mid/right) models. \onetoseq can apply both beam search (mid) and greedy decoding (right).}
    \label{fig:one2one_vs_one2seq}
\end{figure*}

Despite this unique nature, KPG has been essentially ``brute-forced'' into the sequence-to-sequence (Seq2Seq) \citep{sutskever2014seq2seq} framework in the existing literature \citep{meng2017deep,chen2018kp_correlation,ye2018kp_semi,chen2019guided,yuan2018diversekp,chan2019neural,zhao2019incorporating,chen2019integrated}.
The community has approached the unique challenges with much ingenuity in problem formulation, model design, and evaluation. For example, multiple target phrases have been reformulated by either splitting into one phrase per data point or joining into a single sequence with delimiters (Figure~\ref{fig:one2one_vs_one2seq}), both allowing straightforward applications of existing neural techniques such as Seq2Seq. In accordance with the tremendous success and demonstrated effectiveness of neural approaches, steady progress has been made in the past few years --- at least empirically --- across various domains, including sub-areas where it was previously shown to be rather difficult (e.g., in generating keyphrases that are not present in the source text).

Meanwhile, with the myriad of KPG's unique challenges comes an ever-growing collection of studies that, albeit novel and practical, may quickly proliferate and overwhelm. We are therefore motivated to present this study as --- to the best of our knowledge --- the first systematic investigation on such challenges as well as the effect of interplay among their solutions. We hope this study can serve as a practical guide to help researchers to gain a more holistic view on the task, and to profit from the empirical results of our investigations on a variety of topics in KPG including model design, evaluation, and hyper-parameter selection. 


The rest of the paper is organized as follows. We first enumerate specific challenges in KPG due to the multiplicity of its target, and describe general setups for the experiments. We subsequently present experimental results and discussions to answer three main questions:\\
1. How well do KPG models generalize to various testing distributions?\\
2. Does the order of target keyphrases matter while training \onetoseq ?\\
3. Are larger training data helpful? How to better make use of them?


\section{Unique Challenges in KPG}
\label{section:background}
Due to the multiplicity of the generation targets, KPG is unique compared to other NLG tasks such as summarization and translation.
In this section, we start from providing background knowledge of the KPG problem setup.
Then we enumerate the unique aspects in KPG model designing and training that we focus on in this work.

\paragraph{Problem Definition}
Formally, the task of keyphrase generation (KPG) is to generate a \emph{set} of keyphrases $\{p_1,\dots,p_n\}$ given a source text $t$ (a sequence of words).
Semantically, these phrases summarize and highlight important information contained in $t$,
while syntactically, each keyphrase may consist of multiple words.
A keyphrase is defined as \emph{present} if it is a sub-string of the source text, or as \emph{absent} otherwise.


\paragraph{Training Paradigms}
To tackle the unique challenge of generating multiple targets, existing neural KPG approaches can be categorized under one of two training paradigms: \onetoone~\citep{meng2017deep} or \onetoseq~\citep{yuan2018diversekp},
both based on the Seq2Seq framework.
Their main difference lies in how target keyphrase multiplicity is handled in constructing data points (Figure~\ref{fig:one2one_vs_one2seq}).

Specifically, with multiple target phrases $\{p_1,\dots,p_n\}$, \onetoone takes one phrase at a time and pairs it with the source text $t$ to form $n$ data points $(t,p_i)_{i=1:n}$.
During training, a model learns a one-to-many mapping from $t$ to $p_i$'s, i.e., the same source string usually has multiple corresponding target strings.
In contrast, \onetoseq concatenates all ground-truth keyphrases $p_i$ into a single string:\\
$P=\bos p_1 \sep \cdots \sep p_n \eos$
(i.e., prefixed with \bos, joint with \sep, and suffixed with \eos),
thus forming a single data point $(t, P)$.
A system is then trained to predict the concatenated sequence $P$ given $t$.
By default, we construct $P$ follow the ordering strategy proposed in \citep{yuan2018diversekp}.
Specifically, we sort present phrases by their first occurrences in source text, and append absent keyphrases at the end.
This ordering is denoted as \presabs in \S\ref{section:does_order_matter}.

\paragraph{Architecture}
In this paper, we adopt the architecture used in both \citet{meng2017deep} and \citet{yuan2018diversekp}, using \rnn to denote it.
\rnn is a GRU-based Seq2Seq model \citep{cho14gru} with a copy mechanism \citep{Gu2016copy} and a coverage mechanism \citep{see17gettothepoint}.
We also consider a more recent architecture, Transformer \citep{vaswani17transformer}, which is widely used in encoder-decoder language generation literature \citep{gehrmann2018_bottomup}.
We replace both the encoder GRU and decoder GRU in \rnn by Transformer blocks, and denote this architecture variant as \transformer.
Both the \rnn and \transformer models can be trained with either the \onetoone or \onetoseq paradigm.


In recent years, a host of auxiliary designs and mechanisms have been proposed and developed based on either \onetoone or \onetoseq (see \S\ref{section:related_work}).
In this study, however, we focus only on the ``vanilla'' version of them and we show that given a set of carefully chosen architectures and training strategies, base models can achieve comparable, if not better performance than state-of-the-art methods.
We assume that KPG systems derived from either \onetoone or \onetoseq model would be affected by these factors of model designing in similar ways.

\begin{table*}[!htbp]
    \centering
    \scriptsize
    \begin{tabular}{cc|cccc|cccc|cccc}
    \hline
    & \multirow{3}{*}{Dataset} & \multicolumn{4}{c|}{Present (\fo)}
    & \multicolumn{4}{c|}{Present (\ften)}
    & \multicolumn{4}{c}{Absent (\rfifty)} \\ 
    & & \multicolumn{2}{c}{\onetoone} & \multicolumn{2}{c|}{\onetoseq} 
    & \multicolumn{2}{c}{\onetoone} & \multicolumn{2}{c|}{\onetoseq} 
    & \multicolumn{2}{c}{\onetoone} & \multicolumn{2}{c}{\onetoseq} \\ 
    & & \rnn & \transformer & \rnn & \transformer & \rnn & \transformer & \rnn & \transformer & \rnn & \transformer & \rnn & \transformer \\
    \hline
    \multirow{3}{*}{$\text{D}_0$} & \kpk 
    & 35.3 & 37.4 & 31.2 & 36.2
    & 27.9 & 28.9 & 26.1 & 29.0
    & 13.1 & 22.1 & 3.2 & 15.0
    \\
    & \krapivin 
    & 35.5 & 33.0 & 33.5 & 36.4
    & 27.0 & 26.4 & 26.9 & 28.1
    & 13.7 & 23.8 & 3.3 & 16.6
    \\
    & $\text{D}_0$ Average 
    & 35.4 & 35.2 & 32.3 & \textbf{36.3}
    & 27.4 & 27.7 & 26.5 & \textbf{28.5}
    & 13.4 & \textbf{23.0} &  3.2 & 15.8
    \\
    \hline
    \multirow{4}{*}{$\text{D}_1$} & \inspec 
    & 33.7 & 32.6 & 38.8 & 36.9
    & 32.5 & 30.8 & 38.7 & 36.6
    &  8.2 &  9.2 &  3.7 &  6.7
    \\
    & \nus 
    & 43.4 & 41.1 & 39.2 & 42.3
    & 35.9 & 36.1 & 36.6 & 37.3
    & 11.2 & 18.9 &  2.9 & 12.5
    \\
    & \semeval
    & 35.2 & 35.1 & 36.2 & 34.8
    & 34.6 & 33.0 & 35.0 & 34.2
    &  6.1 & 18.9 &  1.7 & 12.5
    \\
    & $\text{D}_1$ Average 
    & 37.4 & 36.3 & \textbf{38.1} & 38.0
    & 34.4 & 33.3 & \textbf{36.7} & 36.0
    &  8.5 & \textbf{12.7} &  2.8 & 9.2
    \\
    \hline
    $\text{D}_2$ & \duc 
    & 13.4 & 7.8 & \textbf{15.0} & 11.0
    & 13.7 & 8.4 & \textbf{16.0} & 11.4
    &  0.0 & \textbf{0.2} &  0.0 &  0.0
    \\
    \hline
    All & Average 
    & 32.8 & 31.2 & 32.3 & \textbf{32.9}
    & 28.6 & 27.3 & \textbf{29.9} & 29.4
    &  8.7 & \textbf{14.0} &  2.5 &  9.8
    \\
    \hline
  \end{tabular}
  \caption{Testing scores across different model architectures, training paradigms, and datasets. In which, $\text{D}_0$: in-distribution; $\text{D}_1$: out-of-distribution, and $\text{D}_2$: out-of-domain. We provide the average score over each category.} 
  \label{tab:performance-comparison}
\end{table*}

\paragraph{Decoding Strategies}
KPG is distinct from other NLG tasks since it expects a \emph{set} of multi-word phrases (rather than a single sequence) as model predictions.
Depending on the preference of potential downstream tasks, a KPG system can utilize different decoding strategies.
For applications that favor \emph{high recall} (e.g., generating indexing terms for retrieval systems), a common practice is to utilize beam search and take predictions from \emph{all} beams\footnote{This is in contrast to only taking the \textit{single} top beam as in typical NLG tasks.}.
This is applicable in both \onetoone- and \onetoseq-based models to proliferate the number of predicted phrases at inference time.
In this work, we use a beam width of 200 and 50 for \onetoone and \onetoseq, respectively.
On the contrary, some other applications favor \emph{high precision} and small number of predictions (e.g., KG construction), a \onetoseq-based model is capable of decoding greedily, thanks to its nature of generating multiple keyphrases in a sequential manner.

As an example, we illustrate the two decoding strategies in Figure~\ref{fig:one2one_vs_one2seq}.
Specifically, a \onetoone model typically collects output keyphrases from all beams and use the top $k$ phrases as the model output ($k=5$ in the example).
In \onetoseq, either beam search or greedy decoding can be applied. 
For beam search, we use both the order of phrases within a beam and the rankings of beams to rank the outputs.
In the shown example, top 5 beam search outputs are obtained from the 2 beams with highest rankings.
As for greedy decoding, the decoder uses a beam size of 1, and takes all phrases from the single beam as outputs.
In this way, the \onetoseq model can determine the number of phrases to output by itself conditioned on $t$.

\paragraph{Evaluation}
Due to the multiplicity of targets in KPG task, the evaluation protocols are distinct from typical NLG tasks.
A spectrum of evaluation metrics have been used to evaluate KPG systems,
including metrics that truncate model outputs at a fixed number such as \ffive and \ften \citep{meng2017deep}; metrics that evaluate a model's ability of generating variant number of phrases such as \fo and \fm \citep{yuan2018diversekp}; metrics that evaluate absent keyphrases such as Recall@50 (\rfifty).
Detailed definitions of the metrics are provided in Appendix~\ref{appendix:metric}.
Due to space limit, we mainly discuss \fo, \ften and \rfifty in the main content, complete results with all common metrics are included in Appendix~\ref{appendix:complete_results}.
We save model checkpoints for every 5,000 training steps and report test performance using checkpoints that produce the best \fo or \rfifty on the \kpk validation set. 


\paragraph{Datasets}
A collection of datasets in the domain of scientific publication (\kpk, \inspec, \krapivin, \nus, and \semeval) and news articles (\duc) have been widely used to evaluate KPG task.
Following previous work, we train models using the training set of \kpk since its size is sufficient to support the training of deep neural networks. Evaluation is performed on \kpk's test set as well as all other datasets without fine-tuning.
Details of the datasets are shown in Appendix~\ref{appendix:stats}.

\section{Generalizability}
\label{section:generalization}

In this section, we show and analyze the generalization performance of KPG systems from 2 dimensions: model architecture and training paradigm.
Specifically, we compare the two model architectures (i.e., \rnn and \transformer) as described in \S\ref{section:background}.
For each model architecture, we train the KPG model using either of the training paradigms (i.e., \onetoone or \onetoseq) also as described in \S\ref{section:background}.

To better understand model variants' generalization properties, we categorize the 6 testing sets into 3 classes according to their distribution similarity with the training data (\kpk), as shown in Table~\ref{tab:performance-comparison}.
Concretely, \kpk and \krapivin are in-distribution test sets (denoted as $\text{D}_0$), since they both contain scientific paper abstracts paired with keyphrases provided by their authors.
\inspec, \nus and \semeval are out-of-distribution test sets (denoted as $\text{D}_1$), they share same type of source text with $\text{D}_0$, but with additionally labeled keywords by third-party annotators.
\duc is a special test set which uses news articles as its source text.
Because it shares the least domain knowledge and vocabulary with all the other test sets, we call it out-of-domain test set (denoted as $\text{D}_2$).


\paragraph{Model Architecture: \rnn vs \transformer}
The first thing to notice is that on present KPG, the models show consistent trends between \ften and \fo.
We observe that \transformer models significantly outperform \rnn models when trained with the \onetoseq paradigm on $\text{D}_0$ test sets.
However, when test data distribution shift increases, on $\text{D}_1$ test sets, \rnn models starts to outperform \transformer; eventually, when dealing with $\text{D}_2$ test set, \rnn outperforms \transformer by a large margin.
On models trained with \onetoone paradigm, we observe a similar trend.
On $\text{D}_0$ data, \transformer models achieve comparable \ften and \fo scores with \rnn, when data distribution shift increases, \rnn models produce better results.

On the contrary, for absent KPG, \transformer outperforms \rnn by a significant margin in all experiment settings.
This is especially obvious when models are trained with \onetoseq paradigm, where \rnn models barely generalize to any of the testing data and produce an average \rfifty of 2.5.
In the same setting, \transformer models get an average \rfifty of 9.8, which is $4\times$ higher than \rnn.

To further study the different generation behaviors between \rnn and \transformer, we investigate the average number of unique predictions generated by either of the models.
As shown in Figure~\ref{fig:order_unique_all} in Appendix~\ref{appendix:does_order_matter}, comparing results of order \presabs in sub-figure a/b (\rnn) with sub-figure c/d (\transformer), we observe that \transformer is consistently generating more unique predictions than \rnn, in both cases of greedy decoding (4.5 vs 4.2) and beam search (123.3 vs 96.8).
We suspect that generating a more diverse set of keyphrases may have a stronger effect on in-distribution test data. 
The generated outputs during inference are likely to represent the distribution learned from the training data, when the test data share the same (or similar) distribution, a larger set of unique predictions leads to a higher recall --- which further contributes to their F-scores.
In contrast, on test sets which data distribution is far from training distribution, the extra predictions may not be as useful, and even hurts precision.
Similarly, because we evaluate absent KPG by the models' recall, \transformer models --- produce more unique predictions --- can always outperform \rnn models.\footnote{Our \transformer and \rnn models follow \citet{vaswani17transformer} and \citet{meng2017deep}'s hyper-parameter settings respectively. 
\rnn is significantly lighter than \transformer.
We conduct experiments with a much larger \rnn but only observe marginal performance boost against \citet{meng2017deep}'s setting.}

\begin{figure}
    \centering
    \includegraphics[width=0.5\textwidth]{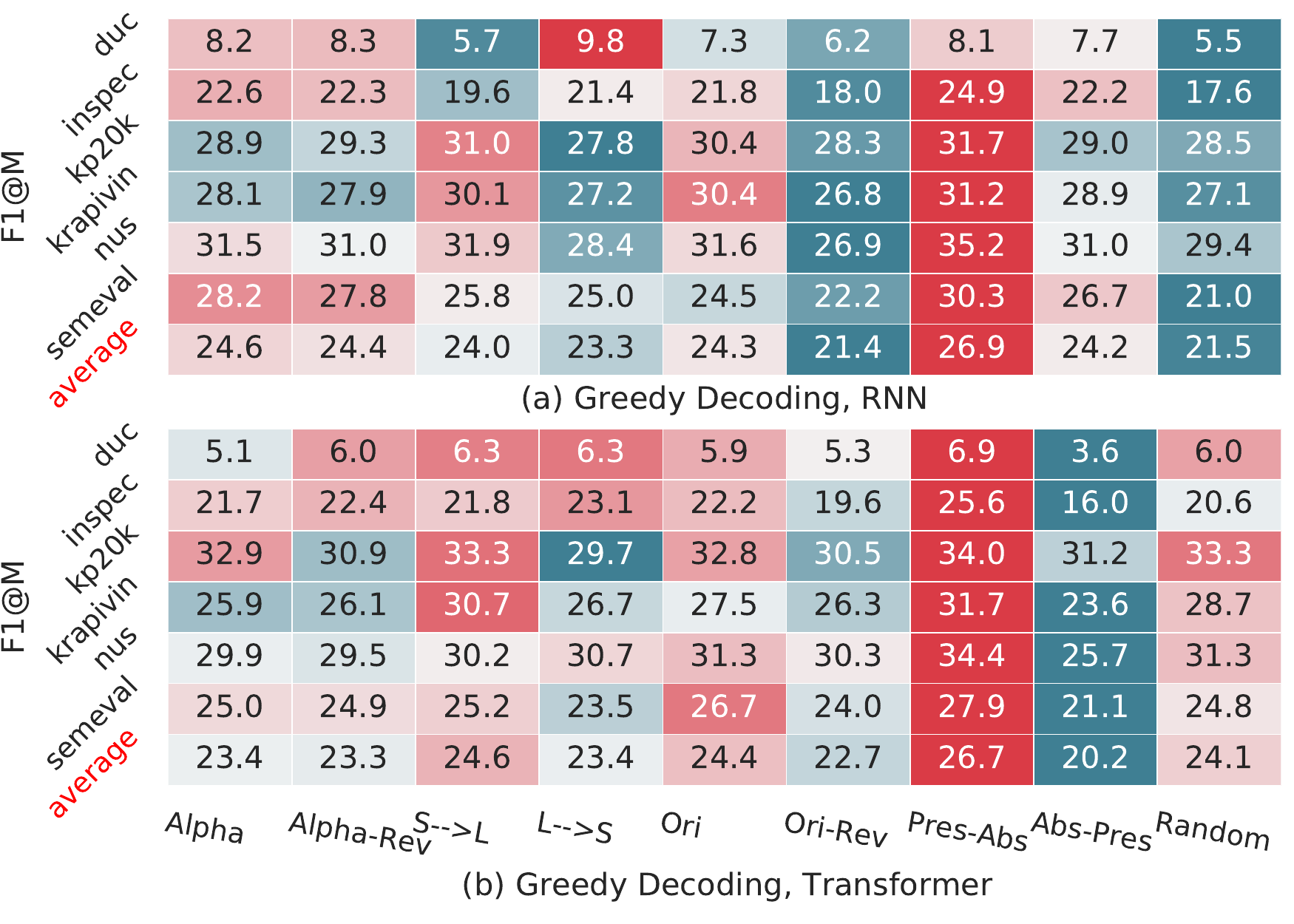}
    \caption{Present KPG testing scores (\fm). Colors represent the relative performance, normalized per row.}
    \label{fig:order_present_fm}
\end{figure}

\paragraph{Training Paradigm: \onetoone vs \onetoseq}
We observe that on present KPG tasks, models trained with the \onetoseq paradigm outperforms \onetoone in most settings, this is particularly clear on $\text{D}_1$ and $\text{D}_2$ test sets.
We believe this is potentially due to the unique design of the \onetoseq training paradigm where at every generation step, the model conditions its decision making on all previously generated tokens (phrases).
Compared to the \onetoone paradigm where multiple phrases can only be generated independently by beam search in parallel, the \onetoseq paradigm can model the dependencies among tokens and the dependencies among phrases more explicitly.

\begin{figure*}[t!]
    \centering
    \includegraphics[width=0.8\textwidth]{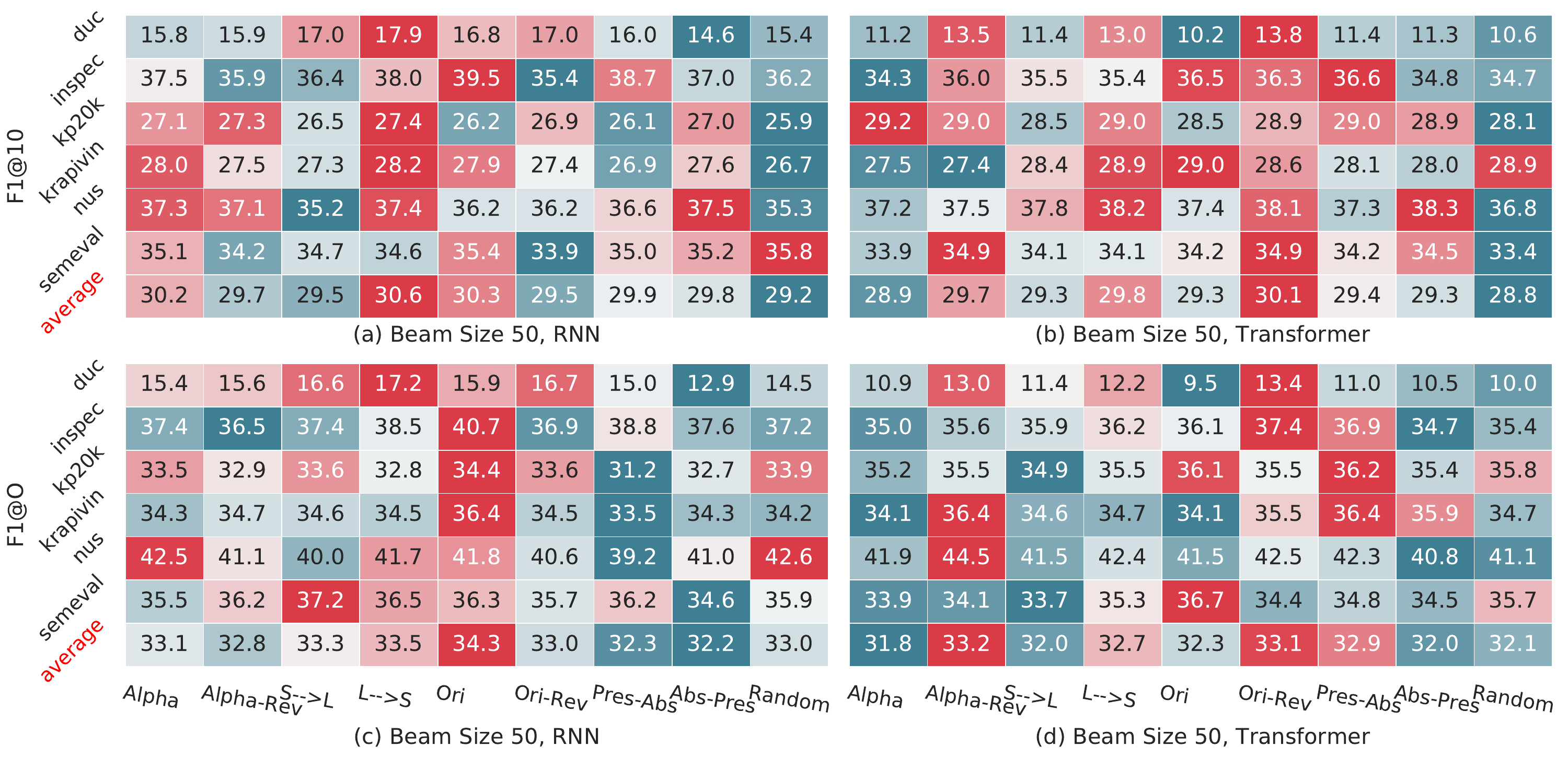}
    \caption{Present KPG testing scores (\textbf{top:} \ften, \textbf{bottom:} \fo). Colors represent the relative performance, normalized per row.}
    \label{fig:order_present_f10o}
\end{figure*}

However, on absent KPG, \onetoone consistently outperforms \onetoseq.
Furthermore, only when trained with \onetoone paradigm, an \rnn-based model can achieve \rfifty scores close to \transformer-based models.
This may because a \onetoseq model tends to produce more duplicated predictions during beam search inference.
By design, every beam is a string that contains multiple phrases that concatenated by the delimiter \sep, there is no guarantee that the phrase will not appear in multiple beams.
In the example shown in Figure~\ref{fig:one2one_vs_one2seq}, ``topic tracking'' is such a duplicate prediction that appears in multiple beams. In fact, the proportion of duplicates in \onetoseq predictions is more than 90\%.
This is in contrast with beam search on \onetoone models, where each beam only contains a single keyphrase thus has a much lower probability of generating duplication.\footnote{Due to post-processing such as stemming, \onetoone model may still produce duplication.}

\section{Does Order Matter in \onetoseq?}
\label{section:does_order_matter}
In the \onetoone paradigm (as shown in Figure~\ref{fig:one2one_vs_one2seq}), each data example is split to multiple equally weighted data pairs, thus it generates phrases without any prior on the order. 
In contrast, \onetoseq training has the unique capability of generating a varying number of keyphrases in a single sequence. 
This inductive bias enables a model to learn dependencies among keyphrases, and also to implicitly estimate the number of target phrases conditioned on the source text.
However, the \onetoseq approach introduces a new complication.
During training, the Seq2Seq decoder takes the concatenation of multiple target keyphrases as target.
As pointed out by \citet{vinyals2016_order_matters}, order matters in sequence modeling tasks;
yet the ordering among the target keyphrases has not been fully investigated and its effect to the models' performance remains unclear.
Several studies have noted this problem \citep{ye2018kp_semi,yuan2018diversekp} without further exploration.

\paragraph{Ordering Definition}
To explore along this direction, we first define nine ordering strategies for concatenating target phrases.\\
\begin{itemize}[leftmargin=*]
    \item \random: Randomly shuffle the target phrases. Because of the \textit{set generation} nature of KPG, we expect randomly shuffled target sequences help to learn an order-invariant decoder.
    \item \ori: Keep phrases in their original order in the data (e.g., provided by the authors of source texts). This was used by~\citet{ye2018kp_semi}.
    \item \orirev: Reversed order of \ori.
    \item \stol: Phrases sorted by lengths (number of tokens, from short to long).
    \item \ltos: Reversed order of \stol.
    \item \alphab: Sort phrases by alphabetical order.
    \item \alpharev: Reversed order of \alphab.
    \item \presabs: Sort present phrases by their first occurrences in source text. Absent phrases are shuffled and appended to the end of the present phrase sequence. This was used by \citep{yuan2018diversekp}.
    \item \abspres:  Similar to \presabs, but prepending absent phrases to the beginning.
\end{itemize}



\paragraph{Greedy Decoding}
In Figure~\ref{fig:order_present_fm}, we show the \rnn and \transformer model's \fm on present KPG task, equipped with greedy decoding.
In this setting, the model simply chooses the token with the highest probability at every step, and terminates either upon generating the \eos token or reaching the maximum target length limit (40). 
This means the model predicts phrases solely relying on its innate distribution learned from the training data, and thus this performance could somewhat reflect to which degree the model fits the training distribution and understands the task.

Through this set of experiments, we first observe that each model demonstrates consistent performance across all six test datasets, indicating that ordering strategies play critical roles in training \onetoseq models when greedy decoding is applied.
When using the \rnn architecture, \random consistently yields lower \fm than other ordering strategies on all datasets.
This suggests that a consistent order of the keyphrases is beneficial.
However, \transformer models show a better resistance against randomly shuffled keyphrases and produce average tier performance with the \random ordering. 
Meanwhile, we observe that \presabs outperforms other ordering strategies by significant margins.
A possible explanation is that with this order (of occurrences in the source text),
the current target phrase is always to the right of the previous one,
which can serve as an effective prior for the attention mechanism throughout the \onetoseq decoding process.
We observe similar trends in greedy decoding models' \fo and \ften, due to space limit, we refer readers to Figure~\ref{fig:order_present_fo},~\ref{fig:order_present_f10} in Appendix~\ref{appendix:does_order_matter}.

\paragraph{Beam Search}
Next, we show results obtained from the same set of models equipped with beam search (beam width is 50) in Figure~\ref{fig:order_present_f10o} (a/b).
Compared with greedy decoding (Figure~\ref{fig:order_present_f10}, Appendix~\ref{appendix:does_order_matter}), we can clearly observe the overall \ften scores have positive correlation with the beam width (greedy decoding is a special case where beam width equals to 1).
We observe that compared to the greedy decoding case, the pattern among different ordering strategies appears to be less clear, with the scores distributed more evenly across different settings (concretely, the absolute difference between max average score and min average score is lower).

We suspect that the uniformity among different ordering strategies with beam search may be due to the limitation of the evaluation metric \ften.
The metric \ften truncates a model's predictions to 10 top-ranked keyphrases. 
By investigation, we find that during greedy decoding, the number of predictions acts as a dominant factor, this number varies greatly among different ordering.
With greedy decoding, \presabs can generally predict more phrases than the others, which explains its performance advantage (Figure~\ref{fig:order_unique_present} (a/c), Appendix~\ref{appendix:does_order_matter}). 
However, as the beam width increases, all models can predict more than 10 phrases (Figure~\ref{fig:order_unique_present} (b/d), Appendix~\ref{appendix:does_order_matter}).
In this case, the \ften is contributed more by a model' ability of generating more high quality keyphrases within its top-10 outputs, rather than the amount of predictions.
Therefore, the performance gap among ordering strategies is gradually narrowed in beam search.
For instance, we observe that the \ften difference between \presabs and \stol produced by \rnn is 3.5/2.0/1.0/0.2 when beam width is 1/10/25/50. 

To validate our assumption, we further investigate the same set of models' performance on \fo, which strictly truncates the generated keyphrase list by the number of ground-truth keyphrases $\mathcal{O}$ (where in most cases $\mathcal{O} < 10$).
Under this harsher criterion, a model is required to generate more high quality keyphrases within its top-$\mathcal{O}$ outputs.
From Figure~\ref{fig:order_present_f10o} (c/d), we observe that the scores are less uniformly distributed, this indicates a larger difference between different order settings.
Among all orders, \ori produces best average \fo with \rnn, whereas \alpharev and \orirev produce best average \fo with \transformer.

In our curated list of order settings, there are 3 pairs of orderings with reversed relationship (i.e., \stol vs \ltos, \alphab vs \alpharev, \ori vs \orirev).
Interestingly, we observe that when beam search is applied, these orderings often show a non-negligible score difference with their counterparts.
This also suggests that order matters since specific model architecture and training paradigm often has its own preference on the phrase ordering.


It is also worth mentioning that when we manually check the output sequences in test set produced by \alphab ordering, we notice that the model is actually able to retain alphabetical order among the predicted keyphrases, hinting that a Seq2Seq model might be capable of learning simple morphological dependencies even without access to any character-level representations.

\begin{figure}[t!]
    \centering
    \includegraphics[width=0.5\textwidth]{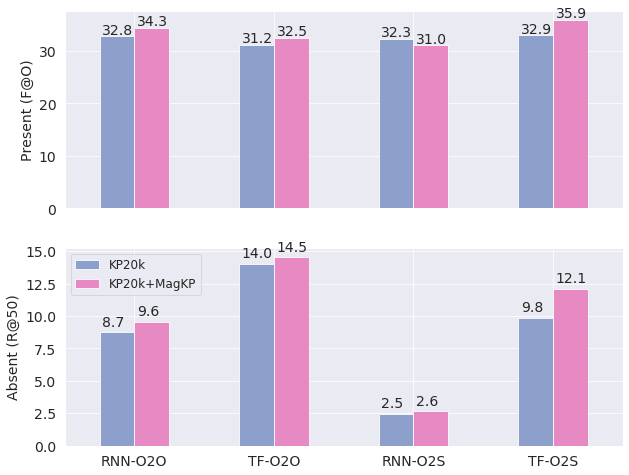}
    \caption{Comparing models trained solely with \kpk against with additional \magkp data. }
    \label{fig:magkp_alternate}
\end{figure}

\paragraph{Ordering in Absent KPG}
We report the performance of the same set of models on the absent portion of data in Figure~\ref{fig:order_absent}, Appendix~\ref{appendix:does_order_matter}.
Although achieving relatively low \rfifty in most settings, scores produced by various orderings show clear distinctions, normalized heat maps suggest that the rankings among different orderings tend to be consistent across all testing datasets.
In general, \presabs produces better absent keyphrases across different model architectures. 
Due to the space limit, we encourage readers to check out Appendix~\ref{appendix:does_order_matter}, which provides an exhaustive set of heat maps including all experiment settings and metrics discussed in this section.

\section{Training with More Data}
\label{section:big_data_big_model}

In this section, we further explore the possibility of improving KPG performance by scaling up the training data. Data size has been shown as one of the most effective factors for training language models~\cite{t5raffel2019exploring, ott2018scaling} but it has yet been discussed in the context of KPG.

\paragraph{MagKP Dataset}
We construct a new dataset, namely \magkp, on the basis of Microsoft Academic Graph~\citep{sinha2015magkp}. 
We filter the original MAG v1 dataset (166 million papers, multiple domains) and only keep papers in \textit{Computer Science} and with at least one keyphrase.
This results in 2.7 million data points ($5\times$ larger than \kpk).
This dataset remains noisy despite the stringent filtering criteria, this is because 1) the data is crawled from the web and 2) some keywords are labeled by automatic systems rather than humans. 
This noisy nature brings many interesting observations. 

\begin{figure}[t!]
    \centering
    \includegraphics[width=0.5\textwidth]{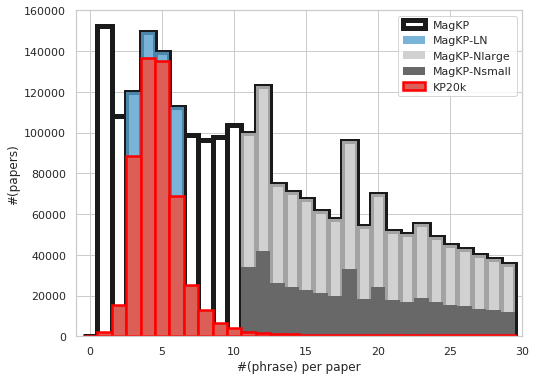}
    \caption{A histogram showing the distribution of \#(kp per document) on \kpk, \magkp and its subsets. Data points with more than 30 keyphrases are truncated.}
    \label{fig:magkp_histgram}
\end{figure}

\paragraph{General Observations}
The first thing we try is to train a KPG model with both \kpk and \magkp.
During training, the two dataset are fed to the model in an alternate manner, we denote this data mixing strategy as \altmix.
In Figure~\ref{fig:magkp_alternate}, we compare models' performance when trained on both \kpk and \magkp against solely on \kpk.
We observe the extra \magkp data brings consistent improvement across most model architecture and training paradigm variants.
This suggests that model KPG models discussed in this work can benefit from additional training data.
Among all the settings, \fo of the \transformerstrip+\onetoseq is boosted by 3 points on present KPG, the resulting score outperforms other variants by a significant margin and even surpass a host of state-of-the-art models (see comparison in Appendix~\ref{appendix:complete_results}).
Again, the same setting obtains a 2.3 boost of \rfifty score on the absent KPG task, makes \transformerstrip+\onetoseq the setting that benefits the most from extra data.
In contrast, the extra \magkp data provide only marginal improvement to \rnn-based models.
On present KPG, \rnnstrip+\onetoseq even has an \fo drop when trained with more data.

As mentioned in \S\ref{section:generalization}, the \rnn model is significantly lighter than \transformer.
To investigate if an RNN with more parameters can benefit more from \magkp, we conduct experiments which use a GRU with much larger hidden size (dubbed \bigrnn).
Results (in Appendix~\ref{appendix:complete_results}) suggest otherwise, extra training data leads to negative effect on \onetoone and only marginal gain on \onetoseq.
We thus believe the architecture difference between \transformer and \rnn is the potential cause, for instance, the built-in self-attention mechanism may help \transformer models learning from noisy data.

\paragraph{Learning with Noisy Data}
To further investigate the performance boost brought by the \magkp dataset on \transformerstrip+\onetoseq, we are curious to know which portion of the noisy data helped the most.
As a naturally way to cluster the \magkp data, we define the noisiness by the number of keyphrases per data point.
As shown in Figure~\ref{fig:magkp_histgram}, the distribution of \magkp (black border) covers a much wider spectrum on the x-axis compared to \kpk (red).
Because keyphrase labels are provided by human authors, a majority of its keyphrase numbers lie in the range of [3, 6]; however, only less than 20\% of the \magkp data overlaps with this number distribution.



We thus break \magkp down into a set of smaller subset: 1) \magkpln is a considerably \textbf{L}ess \textbf{N}oisy subset that contains data points that have 3\texttildelow6 phrases.
2) \magkpnlarge is the \textbf{N}oisy subset in which all data points have more than 10 keyphrases.
3) \magkpnsmall is a randomly sampled subset of \magkpnlarge with the same size as \magkpln.


We also define a set of data mixing strategies to compare against \altmix:
\onlymix: models are trained solely on a single set (or subset) of data;
\mxmix: \kpk and \magkp (or its subset) are split into shards (10k each) and they are randomly sampled during training;
\ftmix: models are pre-trained on \magkp (or its subset) and fine-tuned on \kpk.



\begin{figure}[t!]
    \centering
    \includegraphics[width=0.5\textwidth]{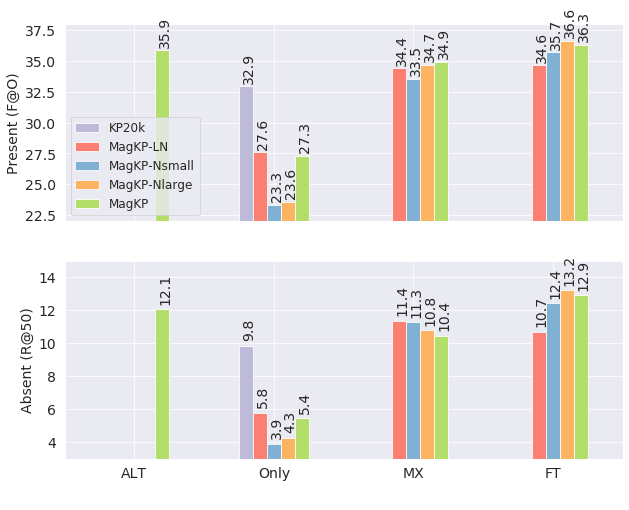}
    \caption{\transformerstrip+\onetoseq trained with \kpk and different subsets of \magkp, using four data mixing strategy. Scores are averaged over all 6 test sets.}
    \label{fig:magkp_finetune}
\end{figure}

In Figure~\ref{fig:magkp_finetune}, we observe that none of the \magkp subsets can match \kpk's performance in the \onlymix~setting.
Because \magkpln and \magkpnsmall share similar data size with \kpk, this suggest the distributional shift between \magkp and the 6 testing sets is significant.
In the \mxmix~setting where \kpk is mixed with noisy data, we observe a notable performance boost compared to \onlymix~(yet still lower than \altmix), however, we do not see clear patterns among the 4 \magkp subsets in this setting.
In the \ftmix~setting, we observe a surge in scores across all \magkp subsets.
In present KPG, both \magkp and \magkpnlarge outperform the score achieved in the \altmix~setting; similarly, in absent KPG, \magkp, \magkpnlarge and \magkpnsmall exceeds the \altmix~score.
This is to our surprise that the subsets considered as noisy provide a greater performance boost, while they perform poorly if ``\onlymix'' trained on these subsets.



To sum up, during our investigation on augmenting \kpk with the noisy \magkp data, we obtain the best performance from a \transformerstrip+\onetoseq model that pre-trained on \magkp and then fine-tuned on \kpk, and this performance has outperformed current state-or-the-art models.
We conjecture that the performance gain may come from data diversity, because \magkp contains a much wider distribution of data compared to the author keyword distribution as in \kpk. 
This inspires us to develop data augmentation techniques to exploit the diversity in unlabeled data. 


\section{Related Work}
\label{section:related_work}

\paragraph{Traditional Keyphrase Extraction}
Keyphrase extraction has been studied extensively for decades. 
A common approach is to formulate it as a two-step process. 
Specifically, a system first heuristically selects a set of candidate phrases from the text using some pre-defined features \citep{witten1999kea,liu2011gap,wang2016ptr,yang2017semisupervisedqa}.
Subsequently, a ranker is used to select the top ranked candidates following various criteria.
The ranker can be bagged decision trees~\citep{medelyan2009human_competitive,lopez2010humb}, Multi-Layer Perceptron, Support Vector Machine ~\citep{lopez2010humb} or PageRank~\citep{Mihalcea2004textrank,le2016unsupervised,wan2008neighborhood_knowledge}.
Compared to the newly developed data driven approaches with deep neural networks, the above approaches suffer from poor performance and the need of dataset-specific heuristic design.

\paragraph{Neural Keyphrase Extraction}
On neural keyphrase extraction task,
\citet{zhang2016twitter,luan2017tagging,gollapalli2017expert_knowledge} use sequence labeling approach;
\citet{subramanian2017kp} use pointer networks to select spans from source text;
\citet{sun2019divgraphpointer} leverage graph neural networks.
Despite improved over tradition approaches, the above methods do not have the capability of predicting absent keyphrases.

\citet{meng2017deep} first propose the CopyRNN model, which both generates words from vocabulary and points to words from the source text --- overcoming the barrier of predicting absent keyphrases.
Following this idea, 
\citet{chen2018kp_correlation, zhao2019incorporating} leverage the attention mechanism to help reducing duplication and improving coverage.
\citet{ye2018kp_semi} propose a semi-supervised training strategy. 
\citet{yuan2018diversekp} propose \onetoseq, which enables a model to generate variable number of keyphrases.  
\citet{chen2019guided,ye2018kp_semi,wang2019topic} propose to leverage extra structure information (e.g., title, topic) to guide the generation.
\citet{chan2019neural} propose an RL model, \citet{swaminathan2020preliminary} propose using GAN for KPG.
\citet{chen2019integrated} retrieve similar documents from training data to help producing more accurate keyphrases.
\citet{chen2020exclusive} introduce hierarchical decoding and exclusion mechanism to prevent from generating duplication.
\citet{ccano2019keyphrase} also propose to utilize more data, but their goal is to bridge KPG with summarization.


\section{Conclusion and Takeaways}
\label{section:conclusion}
We present an empirical study discussing neural KPG models from various aspects.
Through extensive experiments and analysis, we answer the three questions (\S\ref{section:intro}).
Results suggest that given a carefully chosen architecture and training strategy, a base model can perform comparable with fancy SOTA models. Further augmented with (noisy) data in the correct way, a base model can outperform SOTA models (Appendix~\ref{appendix:complete_results}).
We strive to provide a guideline on how to choose such architectures and training strategies, which hopefully can be proven valuable and helpful to the community.

We conclude our discussion with the following takeaways:
\begin{enumerate}
    \item \onetoseq excels at present KPG, while \onetoone performs better on absent KPG. See Section~\ref{section:generalization}.
    \item For present KPG, \transformer performs better on in-distribution data, when distribution or domain shift increase, \rnn can outperform \transformer. See Section~\ref{section:generalization}.
    \item On absent KPG, \transformer is the clear winner. See Section~\ref{section:generalization}.
    \item For \onetoseq, target ordering is important in greedy decoding (with \presabs being an overall good choice). See Section~\ref{section:does_order_matter}.
    \item The effect of target ordering tends to diminish when beam search is performed. See Section~\ref{section:does_order_matter}.
    \item Large and noisy data can benefit KPG. Empirically, a decent way to leverage them is to pre-train on extra data then fine-tune on small in-domain data. See Section~\ref{section:big_data_big_model}.
    \item Copy mechanism helps present prediction while worsening absent performance. See Appendix~\ref{appendix:copy_mechanism}.
    \item Larger beam width is beneficial, especially for absent KPG. However, on present KPG tasks, the benefit is diminished past a certain point and thus computational efficiency needs to be carefully considered. See Appendix~\ref{appendix:beam_search}.
\end{enumerate}

\section*{Acknowledgments}
RM was supported by the Amazon Research Awards for the project ``Transferable, Controllable, Applicable Keyphrase Generation''. 
This research was partially supported by the University of Pittsburgh Center for Research Computing through the resources provided. 
The authors thank the anonymous NAACL reviewers for their helpful feedback and suggestions.

\bibliography{biblio}
\bibliographystyle{acl_natbib}

\clearpage
\appendix

\textbf{\large{Contents in Appendices:}}
\begin{itemize}
    \item In Appendix~\ref{appendix:metric}, we provide the formal definition of all evaluation metrics we used in this work.
    \item In Appendix~\ref{appendix:stats}, we provide detailed statistics of all datasets used in this work.
    \item In Appendix~\ref{appendix:other_remarks}, we provide observation and analysis on additional factors that can affect a KPG system's performance.
    \item In Appendix~\ref{appendix:does_order_matter}, show the set of heat maps that are not shown in the main content due to space limit.
    \item In Appendix~\ref{appendix:complete_results}, we provide a complete set of numbers containing all results discussed in this work, compared with a set of SOTA models in existing literature.
    \item In Appendix~\ref{appendix:implementation_details}, we provide implementation details that helps to reproduce our experiments.
\end{itemize}

\section{Evaluation Metric Definition}
\label{appendix:metric}
In this section, we provide definition of the metrics we use in this work.
All metrics are adopted from \citep{meng2017deep} and \citep{yuan2018diversekp}. To make the results easy to reproduce, we simply \textit{report macro-average scores over all the data examples} in a dataset (rather than removing examples that contain no present/absent phrases). Since some data examples contain no valid present/absent phrase and lead to zero scores, this causes \textit{our results can be lower than previously reported results}.

Given a data example consisting a source text~$\mathcal{X}$ and a list of target keyphrases~$\mathcal{Y}$, suppose that a model predicts a list of unique keyphrases $\mathcal{\hat Y}=(\hat y_1,\dots,\hat y_m)$ ordered by the quality of the predictions $\hat y_i$, and that the ground truth keyphrases for the given source text is the oracle set $\mathcal{Y}$. When only the top $k$ predictions $\mathcal{\hat Y}_{:k}=(\hat y_1,\dots,\hat y_{\min(k,m)})$ are used for evaluation, \textit{precision}, \textit{recall}, and \textit{F$_1$}-score are consequently conditioned on $k$ and defined as:

\begin{equation}
\begin{aligned}
    \pk &= \frac{|\mathcal{\hat Y}_{:k}\cap\mathcal{Y}|}{|\mathcal{\hat Y}_{:k}|}, \:\:\:\:\:\:\:\:
    \rk = \frac{|\mathcal{\hat Y}_{:k}\cap\mathcal{Y}|}{|\mathcal{Y}|},   \\
    \fk &= \frac{2 * \pk * \rk}{\pk + \rk}.
\end{aligned}
\end{equation}

Thus the metrics are defined as:
\begin{itemize}
    \item \ffive: \fk when $k=5$.
    \item \ften: \fk when $k=10$. 
    \item \fo: \textbf{$\mathcal{O}$} denotes the number of oracle (ground truth) keyphrases. In this case, $k=|\mathcal{Y}|$, which means for each data example, the number of predicted phrases taken for evaluation is the same as the number of ground-truth keyphrases.
    \item \fm: \textbf{$\mathcal{M}$} denotes the number of predicted keyphrases. In this case, $k=|\mathcal{\hat Y}|$ and we simply take all the predicted phrases for evaluation without truncation.
    \item \rfifty: \rk when $k=50$.
\end{itemize}

\section{Statistics of Datasets}
\label{appendix:stats}
We provide details of datasets used in this work. 
We use \kpktrain~\cite{meng2017deep} and \magkp~\cite{sinha2015magkp} for training keyphrase generation models, both are built on the basis of scientific publications in Computer Science domain. Nevertheless, their distributions are considerably different, e.g. \magkp data contains 3 times more keyphrases on average than \kpk. This is because \kpk is constructed using real author keywords whereas \magkp may contain a vast amount of keyphrases annotated by automatic systems. Detailed statistics are listed in Table~\ref{tab:train_data_stats}. We also leave out certain amount of data points from \kpk for validation and testing (\kpkvalid and \kpktest).

We also utilize five other datasets for evaluation purposes, as shown in Table~\ref{tab:kp_data_stats}. All except \duc come from scientific publications in Computer Science domain.
\krapivin uses keywords provided by the authors as targets, which is the same as \kpk.
\inspec, \nus, and \semeval contain author-assigned keywords and additional keyphrases provided by third-party annotators. 
\duc, different from all above, is a keyphrase dataset based on news articles. Since it represents a rather different distribution from scientific publication datasets, hypothetically, obtaining decent test score on \duc requires extra generalizability.

\begin{table}[h!]
    \centering
    \scriptsize
    \begin{tabular}{r|c|c|c|c}
        \toprule
        Dataset & \#Data & \#kp & \#unique kp & \#(word) per kp \\
        \midrule
        \text{\kpktrain} & 514K & 2.7M & 700K & 1.92\\
        \text{\magkp} & 2.7M & 41.6M & 6.9M & 3.42\\
        \text{\magkpln} & 522K & 2.3M & 579K & 2.73\\
        \text{\magkpnlarge} & 1.5M & 35.5M & 5.8M & 3.38\\
        \text{\magkpnsmall} & 522K & 12.2M & 2.2M & 3.37\\
        \bottomrule
    \end{tabular}
    \caption{Statistics of training datasets.}
    \label{tab:train_data_stats}
\end{table}

\begin{table}[h!]
    \centering
    \scriptsize
    \begin{tabular}{r|c|c|c}
        \toprule
        Dataset & \#Data & Avg\#kp & \%Pre\\
        \midrule
        \text{\kpk} & $\approx$20K & 5.26 & 63.5\% \\
        \text{\inspec} & 500 & 9.83 & 79.8\% \\
        \text{\krapivin} & 460 & 5.74 & 56.5\% \\
        \text{\nus} & 211 & 11.66 & 51.2\% \\
        \text{\semeval} & 100 & 15.07 & 44.7\% \\
        \text{\duc} & 308 & 8.06 & 97.5\% \\
        \bottomrule
    \end{tabular}
    \caption{Statistics of testing datasets. Avg\#kp indicates the average numbers of target keyphrases, \%Pre denotes percentage of present keyphrases.}
    \label{tab:kp_data_stats}
\end{table}

\section{Other Model Designing Aspects}
\label{appendix:other_remarks}
Besides the findings we discuss in the paper, there exist other important factors affect the general performance of KPG models. We provides two additional empirical results that we think might be of interest to certain readers.

\subsection{Effect of Copy Mechanism}
\label{appendix:copy_mechanism}
Copy mechanism~\cite{Gu2016copy} (also referred to as Pointer Generator~\cite{see17gettothepoint} or Pointer Softmax~\cite{gulcehre2016pointing}) has demonstrated to play a critical role in tasks where texts on the source and target side may overlap, such as summarization~\cite{see17gettothepoint} and keyphrase generation~\cite{meng2017deep}. Basically, it is an additional loss that enables models to extract information from the source side with the help of attentions. Prior studies~\cite{meng2017deep} have shown the importance of copy mechanism with \rnnstrip+\onetoone, but no further comparison has been made. 

\begin{figure}[t!]
    \centering
    \includegraphics[width=0.45\textwidth]{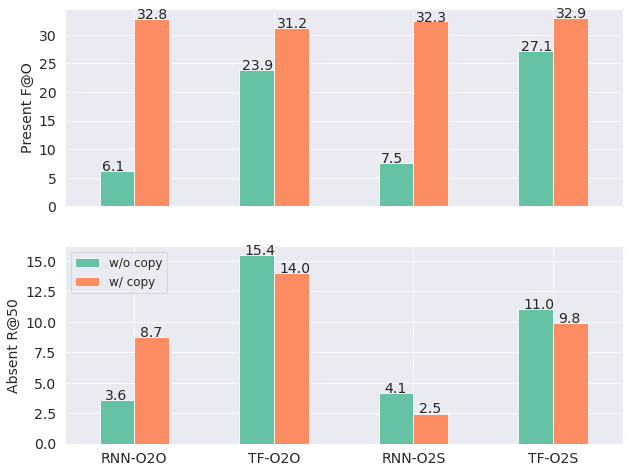}
    \caption{Averaged scores of models with and without utilizing copy mechanism.}
    \label{fig:copy_effect}
\end{figure}

In Figure~\ref{fig:copy_effect}, we present the results of four KPG model variants, equipped with and without copy mechanism. The results show that copy mechanism leads to considerable improvements on present KPG, especially for \rnn. \transformer benefits less from the copy, which may be because its multi-head attentions behave similarly to the copy mechanism even without explicit training losses. With regard to the absent KPG results, copy mechanism only helps \rnnstrip+\onetoone. This suggests that \transformer can achieve consistently better abstractiveness (absent performance) by disabling the copy mechanism at the cost of weaker extractiveness. This dilemma cautions researchers to use copy mechanism more wisely according to specific applications.

\subsection{Effect of Beam Width}
\label{appendix:beam_search}

\begin{figure}[t!]
    \centering
    \includegraphics[width=0.50\textwidth]{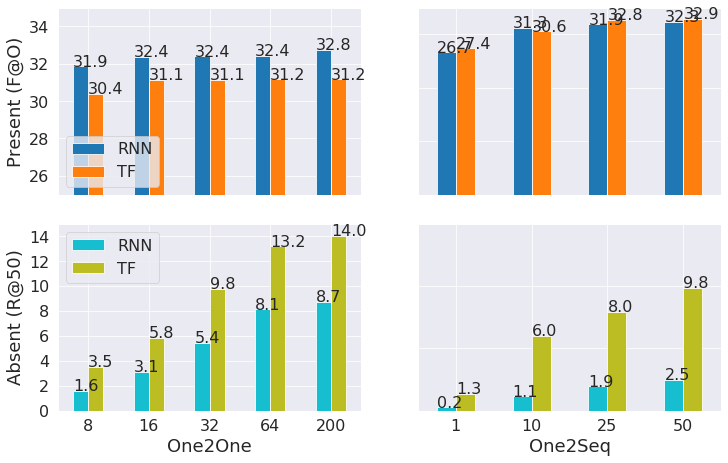}
    \caption{Averaged scores of models using different widths of Beam Search.}
    \label{fig:beam_effect}
\end{figure}

As discussed in \S\ref{section:background}, one unique challenge of the KPG task is due to its multiplicity of target outputs.
As a result, a common strategy is to take multiple beams during decoding in order to obtain more phrases (as opposed to greedy decoding).
This choice is at times not only practical but in fact necessary:
under the \onetoone paradigm, for example, it is crucial to have multiple beams in order to generate multiple keyphrases for a given input.

Generally speaking, KPG and its evaluation metrics are in general favors higher recall.
It is thus not totally unexpected that the high precision scores of greedy decoding are often undermined by notable disadvantages in recall, which in turn leads to losing by large margins in F-scores when compared to results of beam search (with multiple beams).
Empirically, as shown in Figure~\ref{fig:beam_effect}, we observe that beam search can sometimes achieve a relative gain of more than 10\% in present phrase generation , and a much larger performance boost in absent phrase generation, over greedy decoding.

We are also interested in seeing if there exists an optimal beam width.
In Figure~\ref{fig:beam_effect}, we show models' testing performance when various beam widths are used.
In present KPG task with \onetoone (upper left), beam width of 16 already provides an optimal score, larger beam widths (even 200) do not show any further advantage.
Replacing the training paradigm with \onetoseq (upper right), we observe a positive correlation between beam width and testing score --- larger beam widths lead to marginally better testing scores.
However, the improvement (from beam size of 10 to 50) is not significant.

On absent KPG task (lower), both \onetoone and \onetoseq paradigms seem to benefit from larger beam widths.
Testing score shows strong positive correlation with beam width.
We observe that this trend is consistent across different model architectures.

Overall, a larger beam width provides better scores in most settings, but the performance gain diminishes quickly towards very large beam width. 
In addition, it is worth noting that larger beam width also comes with more intense computing demands, for both space and time. 
As an example, in Figure~\ref{fig:beam_effect} (top left), we observe that with the \onetoone training paradigm, a beam width of 200 does not show a significant advantage over 16, however, in terms of computation, beam width of 200 takes about $10 \times$ of the resources compared to 16.
There clearly exists a trade-off between beam width and computational efficiency (e.g., carbon footprint \citep{strubell2019energy}).
We thus hope our results can serve as a reference for researchers, to help them choose beam width more wisely depending on specific tasks.

\section{Does Order Matter in \onetoseq? --- Additional Results}
\label{appendix:does_order_matter}

In \S\ref{section:does_order_matter}, we show models' performance trained with the \onetoseq paradigm using different target ordering strategies. 
Here we provide the complete set of heat maps.

\begin{figure*}[t!]
    \centering
    \includegraphics[width=0.95\textwidth]{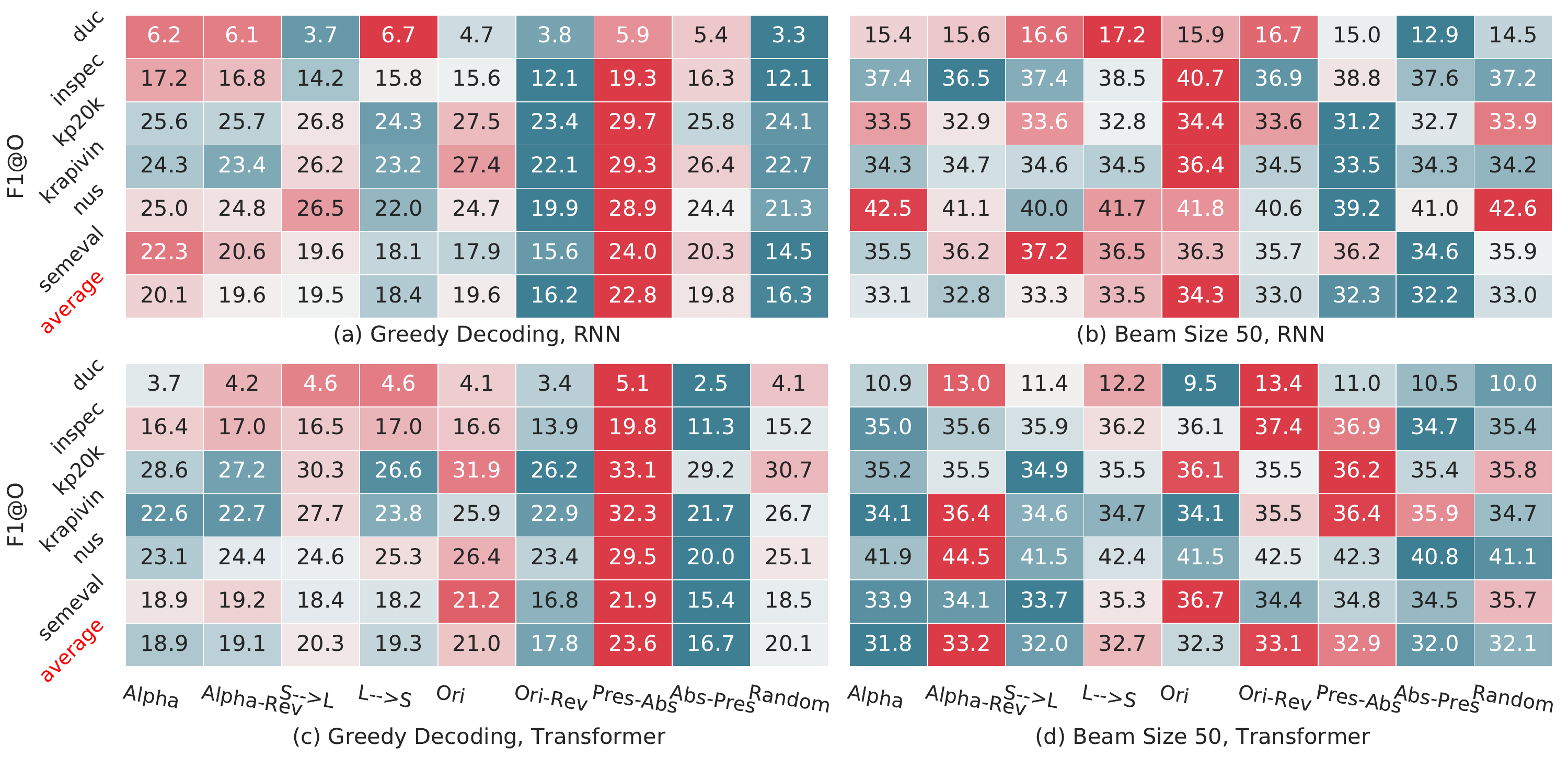}
    \caption{Present keyphrase generation testing scores (\fo). Colors represent the relative performance, normalized per row.}
    \label{fig:order_present_fo}
\end{figure*}

In Figure~\ref{fig:order_present_fo}, we show present KPG testing scores in \fo, when using either greedy decoding or beam search as decoding strategy.

\begin{figure*}[t!]
    \centering
    \includegraphics[width=0.95\textwidth]{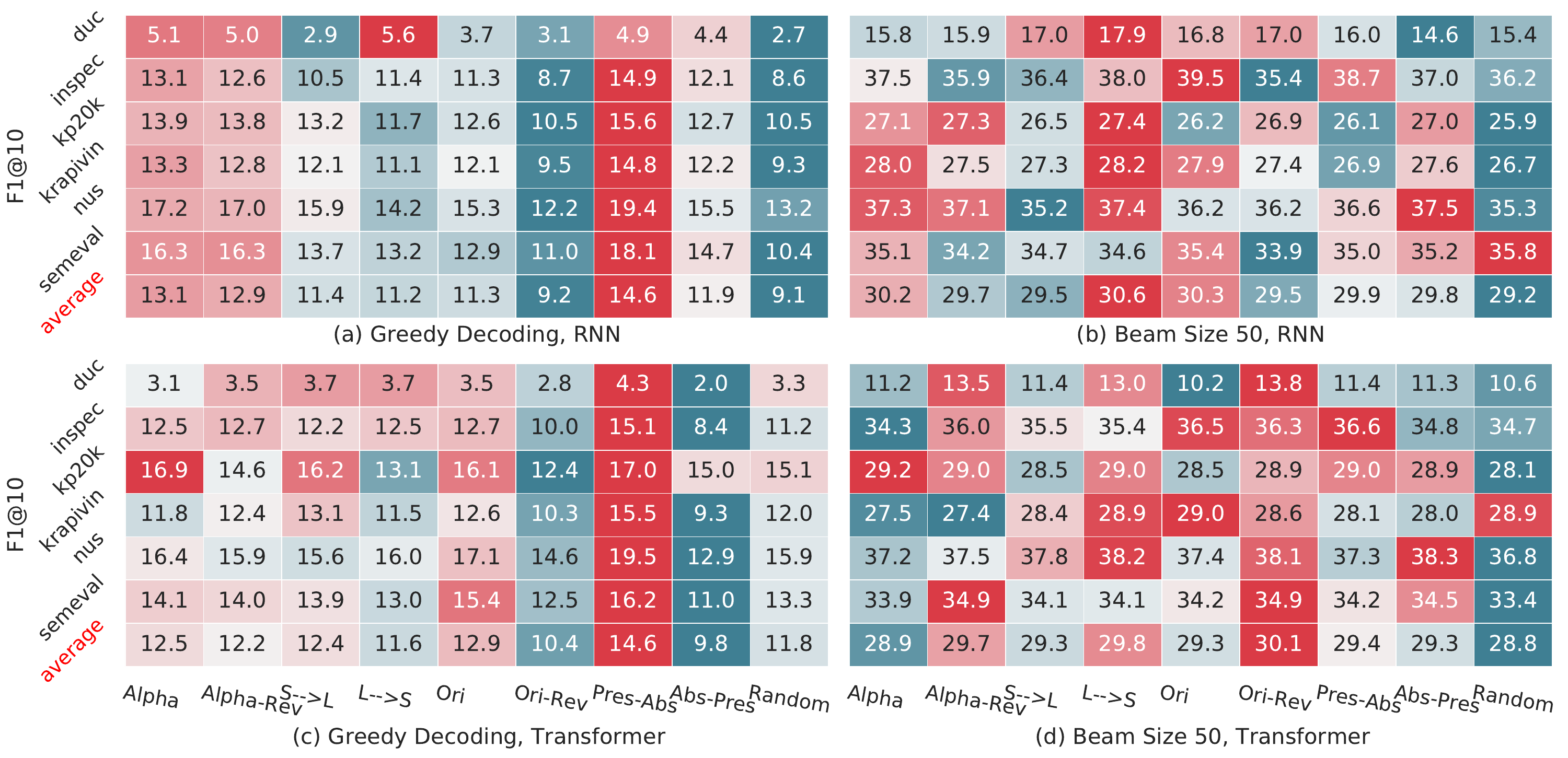}
    \caption{Present keyphrase generation testing scores (\ften). Colors represent the relative performance, normalized per row.}
    \label{fig:order_present_f10}
\end{figure*}

In Figure~\ref{fig:order_present_f10}, we show present KPG testing scores in \ften, when using either greedy decoding or beam search as decoding strategy.

\begin{figure*}[t!]
    \centering
    \includegraphics[width=0.95\textwidth]{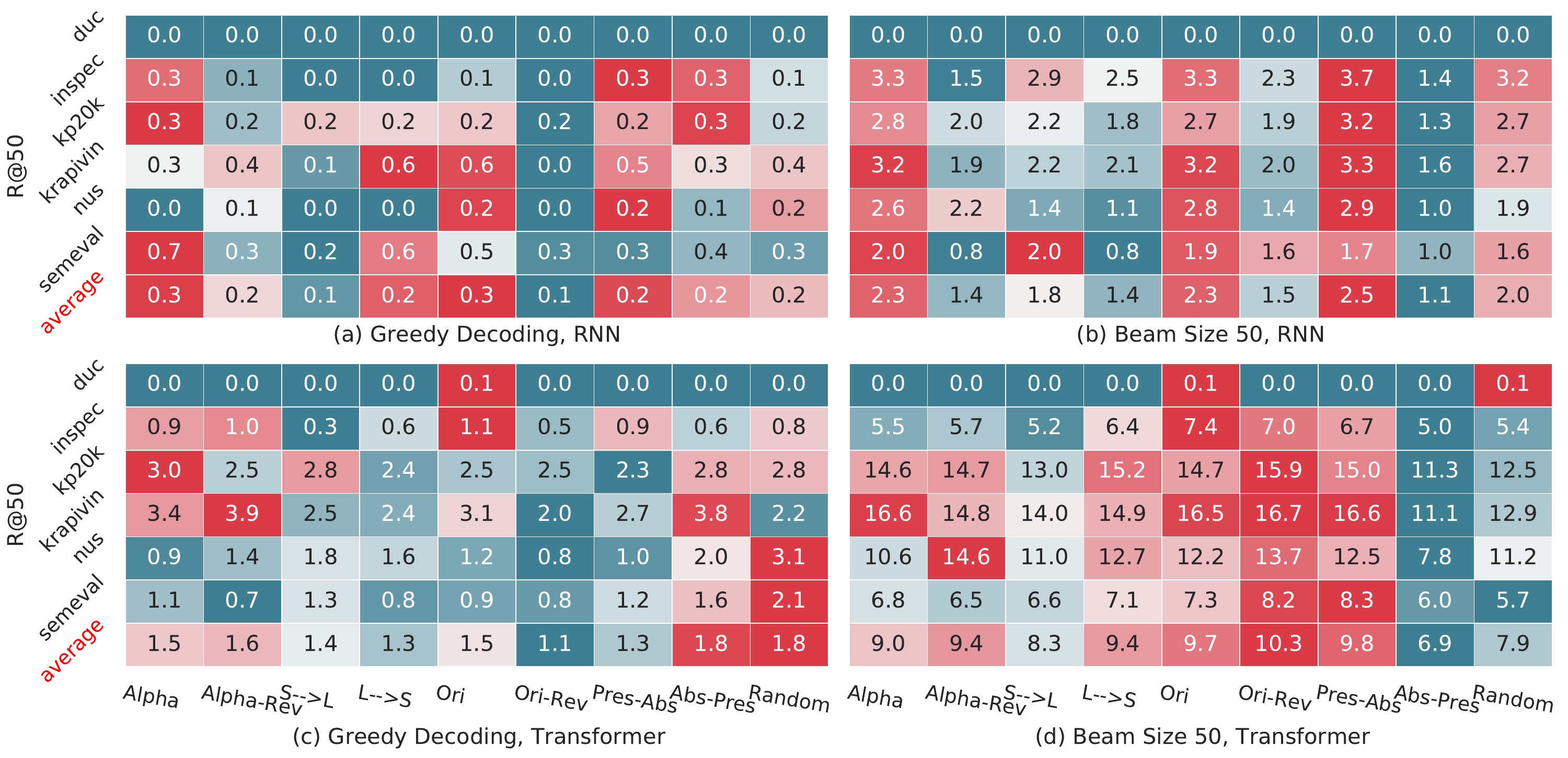}
    \caption{Absent keyphrase generation testing scores on \rfifty. Colors represent the relative performance, normalized per row.}
    \label{fig:order_absent}
\end{figure*}

In Figure~\ref{fig:order_absent}, we show absent KPG testing scores in \rfifty, when using either greedy decoding or beam search as decoding strategy.

\begin{figure*}[t!]
    \centering
    \includegraphics[width=0.95\textwidth]{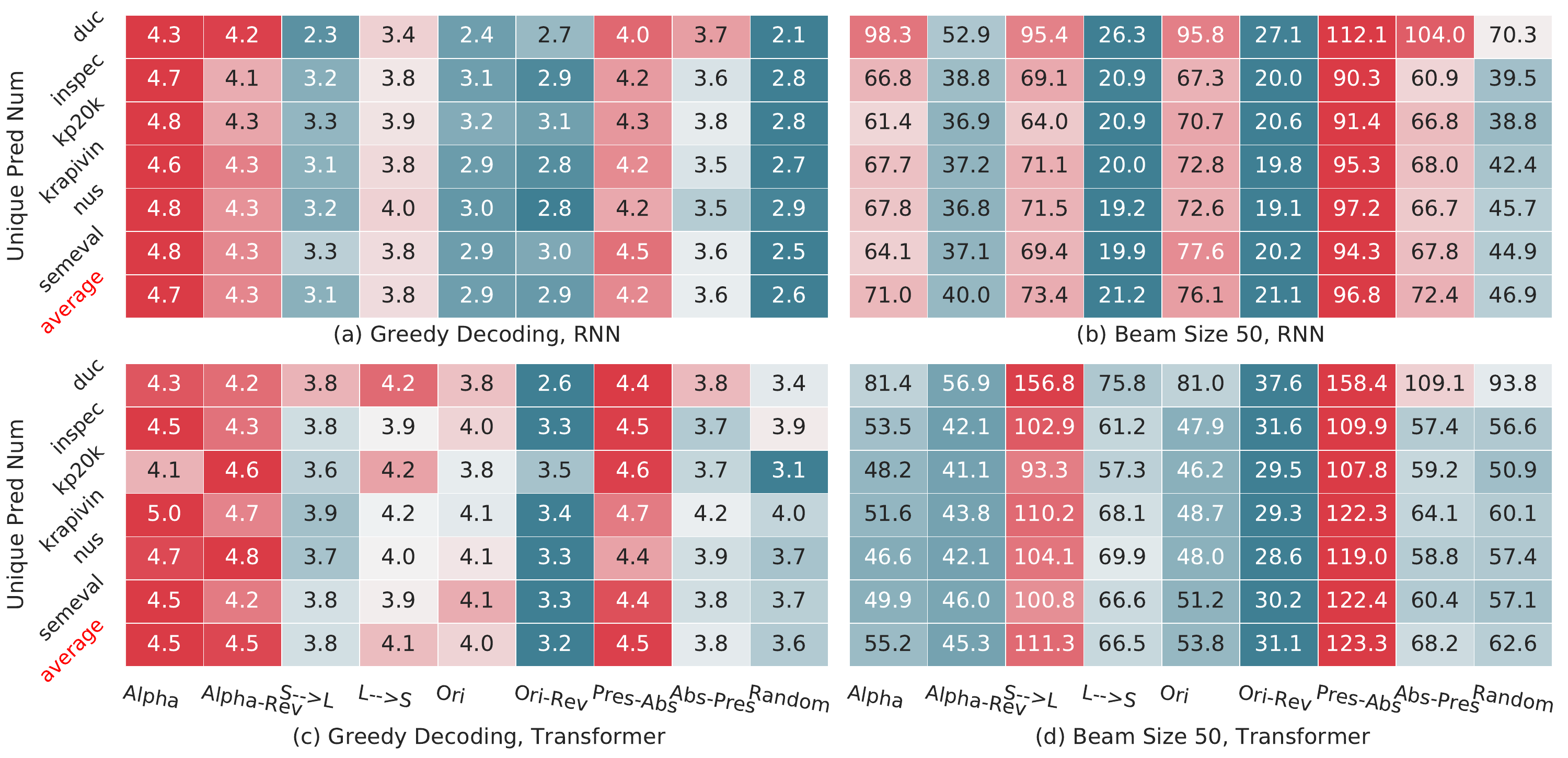}
    \caption{Unique number of keyphrases generated during test. Colors represent the relative performance, normalized per row. Best checkpoints are selected by \fo scores on \kpkvalid.}
    \label{fig:order_unique_all}
\end{figure*}

\begin{figure*}[t!]
    \centering
    \includegraphics[width=0.95\textwidth]{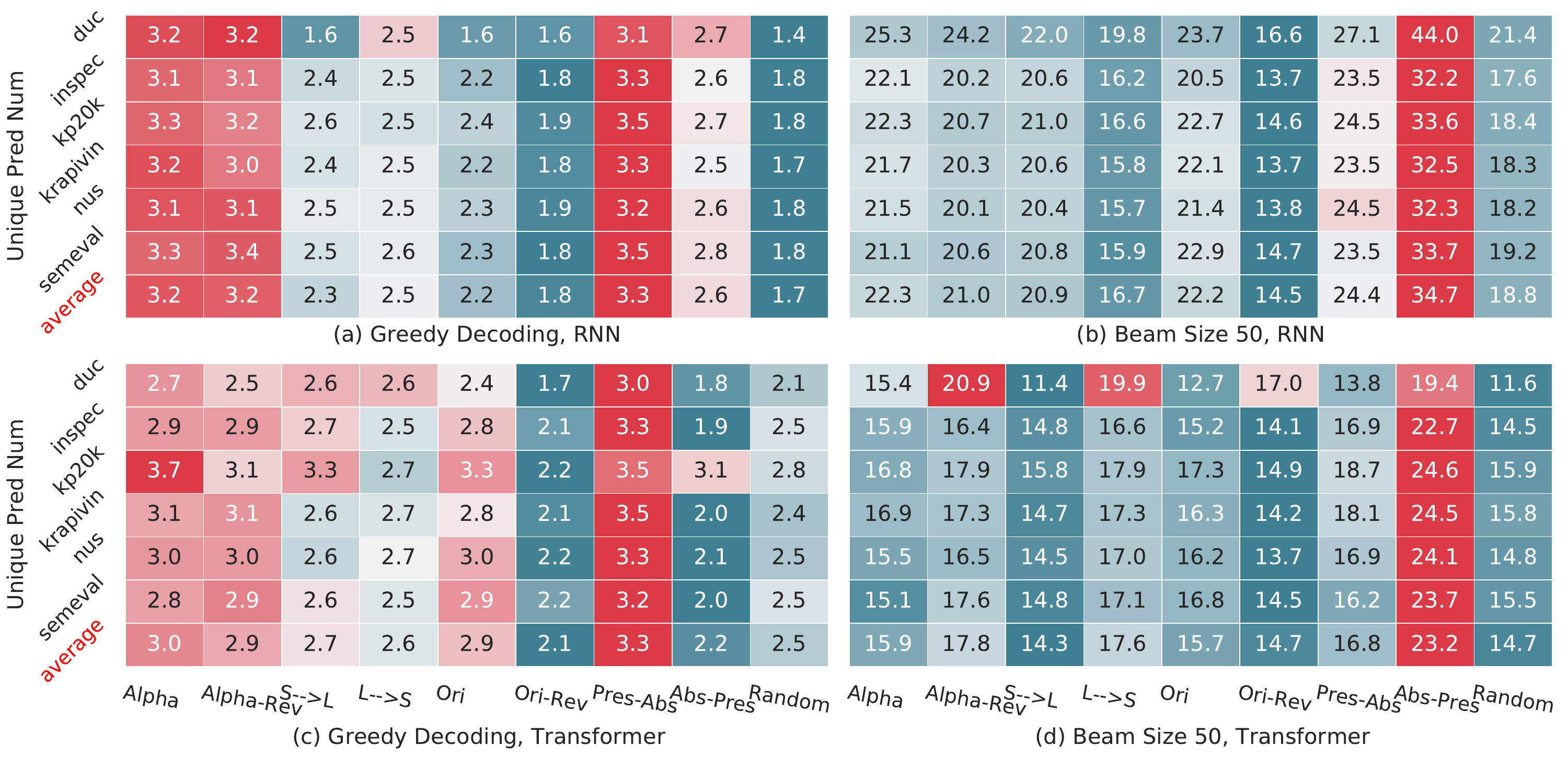}
    \caption{Unique number of present keyphrases generated during test. Colors represent the relative performance, normalized per row. Best checkpoints are selected by \fo scores on \kpkvalid.}
    \label{fig:order_unique_present}
\end{figure*}

\begin{figure*}[t!]
    \centering
    \includegraphics[width=0.95\textwidth]{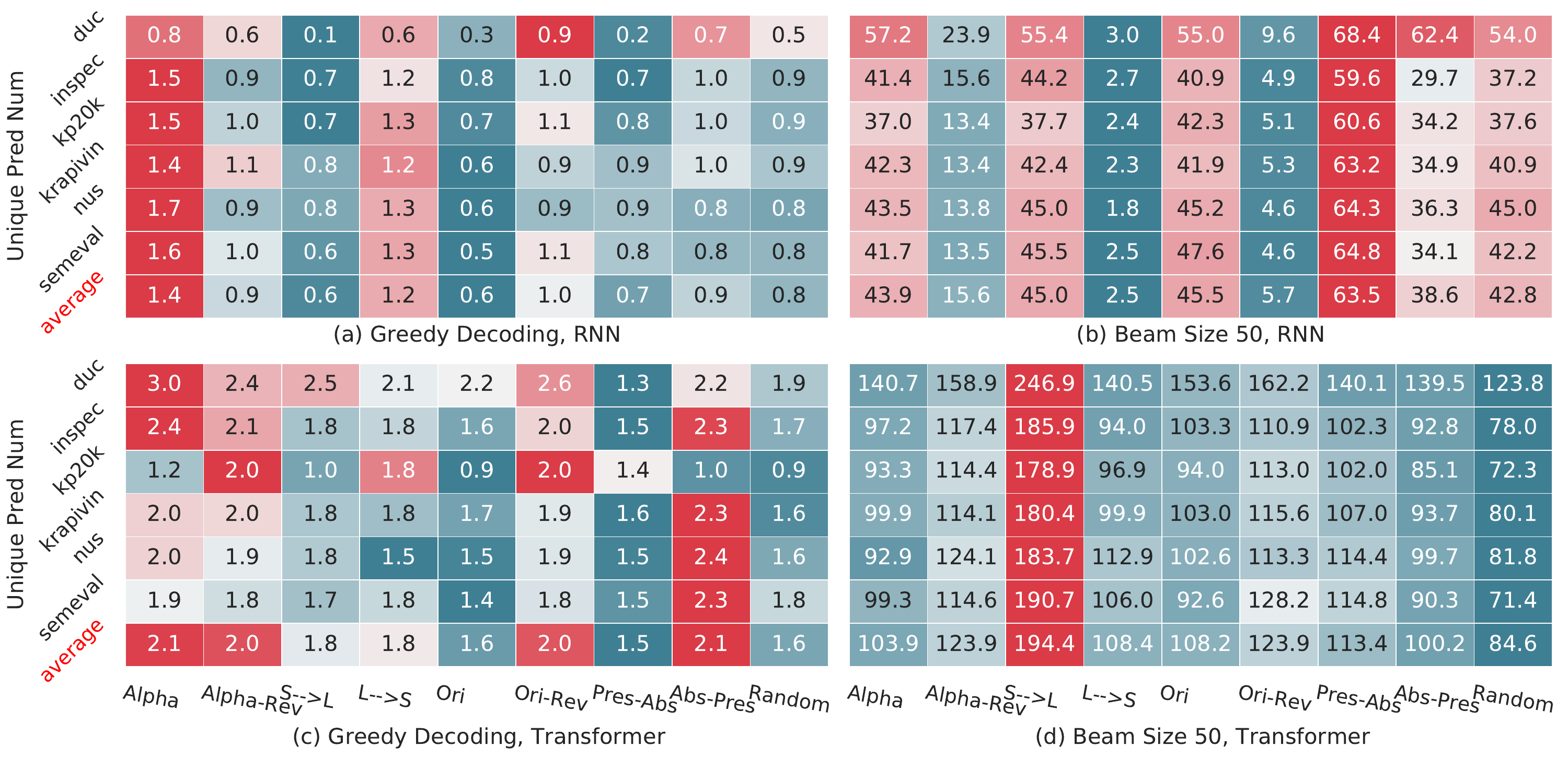}
    \caption{Unique number of absent keyphrases generated during test. Colors represent the relative performance, normalized per row. Best checkpoints are selected by \rfifty scores on \kpkvalid.}
    \label{fig:order_unique_absent}
\end{figure*}

In addition, we shown in Figure~\ref{fig:order_unique_all},~\ref{fig:order_unique_present}, and~\ref{fig:order_unique_absent} the number of unique predictions on all/present/absent KPG tasks.

\section{Complete Results}
\label{appendix:complete_results}

In this section, we report the full set of our experimental results.

In Table~\ref{tab:all_present_scores}, we report all the testing scores on present keyphrase generation tasks.
For all experiments, we use \ffive, \ften, \fo and \fm to evaluate a model's performance.
Additionally, we provide an average score for each of the 4 metrics over all datasets (over each row in Table~\ref{tab:all_present_scores}).

In Table~\ref{tab:all_absent_scores}, we report all the testing scores on absent keyphrase generation tasks.
For all experiments, we use \rten and \rfifty to evaluate a model's performance.
Additionally, we provide an average score for each of the 2 metrics over all datasets (over each row in Table~\ref{tab:all_absent_scores}).

In table~\ref{tab:order_present_scores}, we report detailed present testing scores when model trained with \onetoseq paradigm, using different ordering strategies.
For all experiments, we use \ffive, \ften, \fo and \fm to evaluate a model's performance.

In table~\ref{tab:order_absent_scores}, we report detailed absent testing scores when model trained with \onetoseq paradigm, using different ordering strategies.
For all experiments, we use \rten and \rfifty to evaluate a model's performance.

We also provide scores against all ground-truth phrases (without splitting present/absent) in Table~\ref{tab:all_scores} and~\ref{tab:order_scores} to avoid the inconsistency in data processing (present/absent split may vary by ways of tokenization).

\section{Implementation Details}
\label{appendix:implementation_details}

All the code and data have been released at \url{https://github.com/memray/OpenNMT-kpg-release}, including the new \magkp dataset.

We use the concatenation of title and abstract as the source text. When training with data points contains more than 8 ground-truth keyphrases, we randomly sample 8 from the list to build training target labels. This is to prevent jobs from out-of-memory issues and speed up the training.

We train \rnn models for 100k steps and \transformer for 300k steps. \transformer generally benefits from longer training, especially for absent KPG performance. For the \ftmix~setting, we train models for additional 100k steps.

During the evaluation phase, we replace all punctuation marks with whitespace and tokenize texts/phrases using Python string method \textcolor{blue}{split()}, in order to reduce the errors in phrase matching and present/absent split.

\clearpage

\begin{landscape}
\vfill

\begin{table}
  \label{tab:all_present_scores}
  \centering
  \tiny
  \fontsize{5}{5}\selectfont
\begin{tabular}{c|
  C{0.3cm}C{0.3cm}C{0.3cm}c|
  C{0.3cm}C{0.3cm}C{0.3cm}c|
  C{0.3cm}C{0.3cm}C{0.3cm}c|
  C{0.3cm}C{0.3cm}C{0.4cm}c|
  C{0.3cm}C{0.3cm}C{0.3cm}c|
  C{0.3cm}cC{0.4cm}c|
  C{0.3cm}C{0.3cm}C{0.3cm}c
  }
    \toprule
    \midrule
    & \multicolumn{4}{c|}{\textbf{Average}}
    & \multicolumn{4}{c|}{\textbf{Kp20K}}
    & \multicolumn{4}{c|}{\textbf{Krapivin}}
    & \multicolumn{4}{c|}{\textbf{Inspec}}
    & \multicolumn{4}{c|}{\textbf{NUS}}
    & \multicolumn{4}{c|}{\textbf{SemEval}}
    & \multicolumn{4}{c}{\textbf{DUC}}
    \\
    \midrule
    Model 
    & \noffive & \noften & \nofo & \nofm 
    & \noffive & \noften & \nofo & \nofm 
    & \noffive & \noften & \nofo & \nofm 
    & \noffive & \noften & \nofo & \nofm 
    & \noffive & \noften & \nofo & \nofm 
    & \noffive & \noften & \nofo & \nofm 
    & \noffive & \noften & \nofo & \nofm 
    \\
    \midrule
    \midrule
    \multicolumn{23}{l}{\texttt{\textbf{\onetoone variants}}} \\
    \midrule
    RNN-O2O-KP20k              & 29.7 & 28.6 & 32.8 & 14.9 & 33.1 & 27.9 & 35.3 & 11.1 & 32.0 & 27.0 & 35.5 & 11.2 & 28.5 & 32.5 & 33.7 & 23.1 & 40.2 & 35.9 & 43.4 & 17.7 & 32.9 & 34.6 & 35.2 & 16.4 & 11.4 & 13.7 & 13.4 & 9.7 \\
    RNN-O2O-KP20k-nocopy       & 6.2 & 6.5 & 6.1 & 6.7 & 8.3 & 8.5 & 8.4 & 8.6 & 5.5 & 5.8 & 5.0 & 5.8 & 4.9 & 5.3 & 5.3 & 5.4 & 10.3 & 10.9 & 10.4 & 11.1 & 8.4 & 8.4 & 7.5 & 9.0 & 0.0 & 0.0 & 0.0 & 0.0 \\
    RNN-O2O-KP20k+MagKP-\altmix    & 31.1 & 30.5 & 34.3 & 13.1 & 32.4 & 27.4 & 34.7 & 8.8 & 32.3 & 28.1 & 35.6 & 9.3 & 32.2 & 37.6 & 38.4 & 20.6 & 40.2 & 38.2 & 43.5 & 14.8 & 35.4 & 35.4 & 37.4 & 16.5 & 14.3 & 16.5 & 16.5 & 8.8 \\
\bottomrule
  \end{tabular}
  \caption{Detailed performance (F$_1$-score) of present keyphrase prediction on six datasets. Best checkpoints are selected by present \fo scores on \kpkvalid. \textbf{Boldface}/\underline{Underline} text indicates the best/2nd-best performance in corresponding columns.}
  \label{tab:all_present_scores}
\end{table}
\vfill
\end{landscape}

\begin{landscape}
\vfill

\begin{table}
  \label{tab:all_absent_scores}
  \centering
  \tiny
  \fontsize{6}{6}\selectfont

  \begin{tabular}{c|ccc|ccc|ccc|ccc|ccc|ccc|ccc}
    \toprule
    \midrule
    & \multicolumn{3}{c|}{\textbf{Average}}
    & \multicolumn{3}{c|}{\textbf{Kp20K}}
    & \multicolumn{3}{c|}{\textbf{Inspec}}
    & \multicolumn{3}{c|}{\textbf{Krapivin}}
    & \multicolumn{3}{c|}{\textbf{NUS}}
    & \multicolumn{3}{c}{\textbf{SemEval}}
    & \multicolumn{3}{c}{\textbf{DUC}}
    \\
    \midrule
    Model 
    & \norten & \norfifty & \noratm
    & \norten & \norfifty & \noratm
    & \norten & \norfifty & \noratm
    & \norten & \norfifty & \noratm
    & \norten & \norfifty & \noratm
    & \norten & \norfifty & \noratm
    & \norten & \norfifty & \noratm\\
    \midrule
    \midrule
    \multicolumn{13}{l}{\texttt{\textbf{\onetoone variants}}} \\
    \midrule
    RNN-O2O-KP20k                   & 4.3 & 8.7 & 13.6 & 6.7 & 13.1 & 19.6 & 7.8 & 13.7 & 20.7 & 4.2 & 8.2 & 13.6 & 4.7 & 11.2 & 17.5 & 2.4 & 6.1 & 10.0 & 0.0 & 0.0 & 0.3 \\
    RNN-O2O-KP20k-nocopy            & 1.2 & 3.6 & 6.2 & 2.4 & 5.8 & 9.8 & 1.2 & 4.0 & 8.4 & 1.5 & 2.4 & 3.4 & 1.5 & 5.5 & 9.7 & 0.4 & 3.7 & 6.1 & 0.0 & 0.0 & 0.0 \\
    RNN-O2O-KP20k+MagKP-\altmix     & 5.0 & 9.6 & 14.7 & 6.5 & 13.1 & 19.3 & 8.2 & 16.1 & 23.7 & 5.1 & 9.1 & 15.8 & 7.0 & 13.8 & 18.7 & 3.2 & 5.2 & 10.0 & 0.0 & 0.1 & \textbf{0.8} \\
    \midrule  
    BIGRNN-O2O-KP20k                & 7.4 & 13.9 & 19.2 & 11.4 & 20.2 & 27.8 & 12.7 & 22.5 & 31.7 & 5.6 & 11.3 & 16.8 & 10.1 & 18.6 & 24.8 & 4.3 & \underline{10.1} & 13.7 & 0.1 & \textbf{0.4} & \underline{0.4} \\
\bottomrule
  \end{tabular}
  \caption{Detailed performance (recall scores) of absent keyphrase prediction on six datasets. Best checkpoints are selected by absent \rfifty scores on \kpkvalid. \textbf{Boldface}/\underline{Underline} text indicates the best/2nd-best performance in corresponding columns.}
  \label{tab:all_absent_scores}
\end{table}
\vfill
\end{landscape}

\begin{landscape}
\vfill

\begin{table}
  \label{tab:all_scores}
  \centering
  \tiny
  \fontsize{5}{5}\selectfont
  \begin{tabular}{c|
  C{0.3cm}C{0.3cm}C{0.3cm}c|
  C{0.3cm}C{0.3cm}C{0.3cm}c|
  C{0.3cm}C{0.3cm}C{0.3cm}c|
  C{0.3cm}C{0.3cm}C{0.4cm}c|
  C{0.3cm}C{0.3cm}C{0.3cm}c|
  C{0.3cm}cC{0.4cm}c|
  C{0.3cm}C{0.3cm}C{0.3cm}c
  }
    \toprule
    \midrule
    & \multicolumn{4}{c|}{\textbf{Average}}
    & \multicolumn{4}{c|}{\textbf{Kp20K}}
    & \multicolumn{4}{c|}{\textbf{Krapivin}}
    & \multicolumn{4}{c|}{\textbf{Inspec}}
    & \multicolumn{4}{c|}{\textbf{NUS}}
    & \multicolumn{4}{c|}{\textbf{SemEval}}
    & \multicolumn{4}{c}{\textbf{DUC}}
    \\
    \midrule
    Model 
    & \noffive & \noften & \nofo & \nofm 
    & \noffive & \noften & \nofo & \nofm 
    & \noffive & \noften & \nofo & \nofm 
    & \noffive & \noften & \nofo & \nofm 
    & \noffive & \noften & \nofo & \nofm 
    & \noffive & \noften & \nofo & \nofm 
    & \noffive & \noften & \nofo & \nofm 
    \\ 
    \midrule
    \midrule
    \multicolumn{23}{l}{\texttt{\textbf{\onetoone variants}}} \\
    \midrule
    RNN-O2O-KP20k              & 23.0 & 23.6 & 25.4 & 2.3 & 27.8 & 24.9 & 28.9 & 1.5 & 25.2 & 23.4 & 27.0 & 1.6 & 25.1 & 28.2 & 29.9 & 3.1 & 28.8 & 28.6 & 31.1 & 2.5 & 19.3 & 23.9 & 23.2 & 2.8 & 11.5 & 12.7 & 12.3 & 2.0 \\
    RNN-O2O-KP20k-nocopy       & 1.6 & 1.8 & 1.8 & 0.4 & 2.4 & 2.6 & 2.5 & 0.4 & 2.0 & 2.0 & 1.9 & 0.3 & 1.0 & 1.1 & 1.1 & 0.2 & 2.8 & 3.4 & 3.2 & 0.7 & 1.4 & 1.9 & 1.8 & 0.6 & 0.0 & 0.0 & 0.0 & 0.0 \\
    RNN-O2O-KP20k+MagKP-\altmix    & 24.2 & 25.2 & 27.0 & 2.5 & 27.2 & 24.5 & 28.3 & 1.6 & 26.2 & 24.3 & 27.9 & 1.7 & 27.6 & 32.8 & 34.2 & 3.4 & 28.9 & 29.7 & 31.0 & 2.6 & 21.3 & 23.8 & 24.4 & 2.9 & 13.8 & 16.1 & 16.2 & 2.5 \\
    \midrule
    BIGRNN-O2O-KP20k           & 24.8 & 25.9 & 27.9 & 2.6 & 29.9 & 26.7 & 31.0 & 1.7 & 27.9 & 26.1 & 29.9 & 1.9 & 27.1 & 32.0 & 33.8 & 3.5 & 30.0 & 30.6 & 33.3 & 2.8 & 20.7 & 24.5 & 24.8 & 3.2 & 13.1 & 15.5 & 14.5 & 2.6 \\
\bottomrule
  \end{tabular}
  \caption{Detailed performance (F$_1$-score) of all (present+absent) keyphrase prediction on six datasets. Best checkpoints are selected by \fo scores on \kpkvalid. \textbf{Boldface}/\underline{Underline} text indicates the best/2nd-best performance in corresponding columns.}
  \label{tab:all_scores}
\end{table}
\vfill
\end{landscape}

\clearpage
\begin{landscape}
\vfill

\begin{table}
  \centering
  \tiny
  \fontsize{5}{5}\selectfont
  \begin{tabular}{c|
  C{0.3cm}C{0.3cm}C{0.3cm}c|
  C{0.3cm}C{0.3cm}C{0.3cm}c|
  C{0.3cm}C{0.3cm}C{0.3cm}c|
  C{0.3cm}C{0.3cm}C{0.4cm}c|
  C{0.3cm}C{0.3cm}C{0.3cm}c|
  C{0.3cm}cC{0.4cm}c|
  C{0.3cm}C{0.3cm}C{0.3cm}c
  }
    \toprule
    \midrule
    & \multicolumn{4}{c|}{\textbf{Average}}
    & \multicolumn{4}{c|}{\textbf{Kp20K}}
    & \multicolumn{4}{c|}{\textbf{Krapivin}}
    & \multicolumn{4}{c|}{\textbf{Inspec}}
    & \multicolumn{4}{c|}{\textbf{NUS}}
    & \multicolumn{4}{c|}{\textbf{SemEval}}
    & \multicolumn{4}{c}{\textbf{DUC}}
    \\
    \midrule
    Model 
    & \noffive & \noften & \nofo & \nofm 
    & \noffive & \noften & \nofo & \nofm 
    & \noffive & \noften & \nofo & \nofm 
    & \noffive & \noften & \nofo & \nofm 
    & \noffive & \noften & \nofo & \nofm 
    & \noffive & \noften & \nofo & \nofm 
    & \noffive & \noften & \nofo & \nofm 
    \\
    \midrule
    \midrule
    \multicolumn{23}{l}{\texttt{\textbf{\rnn \greedy}}} \\
    \midrule
    \random                & 21.5 & 21.5 & 21.4 & 21.5 & 28.5 & 28.5 & 28.5 & 28.5 & 27.1 & 27.1 & 26.9 & 27.1 & 17.6 & 17.6 & 17.6 & 17.6 & 29.4 & 29.4 & 28.9 & 29.4 & 21.0 & 21.0 & 21.2 & 21.0 & 5.5 & 5.5 & 5.5 & 5.5 \\
    \alphab                & 24.6 & 24.6 & 23.8 & 24.6 & 28.9 & 28.9 & 27.2 & 28.9 & 28.2 & 28.1 & 26.0 & 28.1 & 22.5 & 22.6 & 22.3 & 22.6 & 31.6 & 31.5 & 30.5 & 31.5 & 28.2 & 28.2 & 28.2 & 28.2 & 8.2 & 8.2 & 8.2 & 8.2 \\
    \alpharev              & 24.4 & 24.4 & 23.3 & 24.4 & 29.3 & 29.3 & 27.4 & 29.3 & 27.8 & 27.9 & 25.5 & 27.9 & 22.3 & 22.3 & 21.9 & 22.3 & 31.0 & 31.0 & 30.6 & 31.0 & 27.8 & 27.8 & 26.4 & 27.8 & 8.2 & 8.3 & 8.3 & 8.3 \\
    \stol                  & 24.0 & 24.0 & 23.8 & 24.0 & 31.0 & 31.0 & 29.7 & 31.0 & 30.1 & 30.1 & 29.2 & 30.1 & 19.6 & 19.6 & 19.4 & 19.6 & 31.9 & 31.9 & 33.0 & 31.9 & 25.8 & 25.8 & 26.0 & 25.8 & 5.7 & 5.7 & 5.7 & 5.7 \\
\bottomrule
  \end{tabular}
  \caption{Detailed present keyphrase prediction performance (F$_1$-score) of \onetoseq trained with different orders. Best checkpoints are selected by present \fo scores on \kpkvalid.}
  \label{tab:order_present_scores}
\end{table}
\vfill
\end{landscape}

\clearpage
\begin{landscape}
\vfill

\begin{table}
  \centering
  \tiny
  \fontsize{6}{6}\selectfont
  
  \begin{tabular}{c|ccc|ccc|ccc|ccc|ccc|ccc|ccc}
    \toprule
    \midrule
    & \multicolumn{3}{c|}{\textbf{Average}}
    & \multicolumn{3}{c|}{\textbf{Kp20K}}
    & \multicolumn{3}{c|}{\textbf{Inspec}}
    & \multicolumn{3}{c|}{\textbf{Krapivin}}
    & \multicolumn{3}{c|}{\textbf{NUS}}
    & \multicolumn{3}{c}{\textbf{SemEval}}
    & \multicolumn{3}{c}{\textbf{DUC}}
    \\
    \midrule
    Model 
    & \norten & \norfifty & \noratm
    & \norten & \norfifty & \noratm
    & \norten & \norfifty & \noratm
    & \norten & \norfifty & \noratm
    & \norten & \norfifty & \noratm
    & \norten & \norfifty & \noratm
    & \norten & \norfifty & \noratm\\
    \midrule
    \midrule
    \multicolumn{13}{l}{\texttt{\textbf{\rnn \greedy}}} \\
    \midrule
    \random                & 0.2 & 0.2 & 0.2 & 0.2 & 0.2 & 0.2 & 0.4 & 0.4 & 0.4 & 0.1 & 0.1 & 0.1 & 0.2 & 0.2 & 0.2 & 0.3 & 0.3 & 0.3 & 0.0 & 0.0 & 0.0 \\
    \alphab                & 0.3 & 0.3 & 0.3 & 0.3 & 0.3 & 0.3 & 0.3 & 0.3 & 0.3 & 0.3 & 0.3 & 0.3 & 0.0 & 0.0 & 0.0 & 0.7 & 0.7 & 0.7 & 0.0 & 0.0 & 0.0 \\
    \alpharev              & 0.2 & 0.2 & 0.2 & 0.2 & 0.2 & 0.2 & 0.4 & 0.4 & 0.4 & 0.1 & 0.1 & 0.1 & 0.1 & 0.1 & 0.1 & 0.3 & 0.3 & 0.3 & 0.0 & 0.0 & 0.0 \\
    \stol                  & 0.1 & 0.1 & 0.1 & 0.2 & 0.2 & 0.2 & 0.1 & 0.1 & 0.1 & 0.0 & 0.0 & 0.0 & 0.0 & 0.0 & 0.0 & 0.2 & 0.2 & 0.2 & 0.0 & 0.0 & 0.0 \\
    \ltos                  & 0.2 & 0.2 & 0.2 & 0.2 & 0.2 & 0.2 & 0.6 & 0.6 & 0.6 & 0.0 & 0.0 & 0.0 & 0.0 & 0.0 & 0.0 & 0.6 & 0.6 & 0.6 & 0.0 & 0.0 & 0.0 \\
\bottomrule
  \end{tabular}
  \caption{Detailed absent keyphrase prediction performance (recall scores) of \onetoseq trained with different orders. Best checkpoints are selected by absent \rfifty scores on \kpkvalid.}
  \label{tab:order_absent_scores}
\end{table}
\vfill
\end{landscape}

\clearpage
\begin{landscape}
\vfill

\begin{table}
  \centering
  \tiny
  \fontsize{5}{5}\selectfont
  \begin{tabular}{c|
  C{0.3cm}C{0.3cm}C{0.3cm}c|
  C{0.3cm}C{0.3cm}C{0.3cm}c|
  C{0.3cm}C{0.3cm}C{0.3cm}c|
  C{0.3cm}C{0.3cm}C{0.4cm}c|
  C{0.3cm}C{0.3cm}C{0.3cm}c|
  C{0.3cm}cC{0.4cm}c|
  C{0.3cm}C{0.3cm}C{0.3cm}c
  }
    \toprule
    \midrule
    & \multicolumn{4}{c|}{\textbf{Average}}
    & \multicolumn{4}{c|}{\textbf{Kp20K}}
    & \multicolumn{4}{c|}{\textbf{Krapivin}}
    & \multicolumn{4}{c|}{\textbf{Inspec}}
    & \multicolumn{4}{c|}{\textbf{NUS}}
    & \multicolumn{4}{c|}{\textbf{SemEval}}
    & \multicolumn{4}{c}{\textbf{DUC}}
    \\
    \midrule
    Model 
    & \noffive & \noften & \nofo & \nofm 
    & \noffive & \noften & \nofo & \nofm 
    & \noffive & \noften & \nofo & \nofm 
    & \noffive & \noften & \nofo & \nofm 
    & \noffive & \noften & \nofo & \nofm 
    & \noffive & \noften & \nofo & \nofm 
    & \noffive & \noften & \nofo & \nofm 
    \\
    \midrule
    \midrule
    \multicolumn{23}{l}{\texttt{\textbf{\rnn \greedy}}} \\
    \midrule
    \random                & 13.9 & 13.9 & 14.0 & 13.9 & 19.3 & 19.3 & 19.6 & 19.3 & 17.5 & 17.4 & 17.6 & 17.4 & 13.9 & 13.9 & 13.9 & 13.9 & 17.5 & 17.5 & 17.5 & 17.5 & 10.3 & 10.3 & 10.3 & 10.3 & 4.9 & 4.9 & 4.9 & 4.9 \\
    \alphab                & 17.0 & 16.5 & 16.9 & 16.5 & 21.3 & 20.4 & 21.6 & 20.4 & 20.1 & 19.6 & 20.2 & 19.6 & 18.2 & 17.7 & 18.0 & 17.7 & 20.1 & 19.4 & 19.8 & 19.4 & 14.9 & 14.5 & 14.5 & 14.5 & 7.5 & 7.5 & 7.5 & 7.5 \\
    \alpharev              & 16.8 & 16.6 & 16.8 & 16.6 & 21.6 & 21.3 & 21.8 & 21.3 & 19.5 & 19.1 & 19.7 & 19.1 & 18.2 & 18.0 & 18.0 & 18.0 & 20.3 & 20.1 & 20.2 & 20.1 & 14.4 & 14.3 & 14.3 & 14.3 & 6.8 & 6.9 & 6.8 & 6.9 \\
    \stol                  & 15.7 & 15.7 & 15.8 & 15.7 & 22.1 & 22.1 & 22.3 & 22.1 & 19.7 & 19.7 & 20.1 & 19.7 & 15.6 & 15.6 & 15.4 & 15.6 & 19.1 & 19.1 & 19.4 & 19.1 & 12.6 & 12.6 & 12.6 & 12.6 & 5.1 & 5.1 & 5.1 & 5.1 \\
\bottomrule
  \end{tabular}
  \caption{Detailed keyphrase prediction performance all phrases (present+absent) of \onetoseq trained with different orders. Best checkpoints are selected by \fo scores on \kpkvalid.}
  \label{tab:order_scores}
\end{table}
\vfill
\end{landscape}

\clearpage

\end{document}